\documentclass[letterpaper]{article} 
\usepackage{aaai24}  
\usepackage{times}  
\usepackage{helvet}  
\usepackage{courier}  
\usepackage[hyphens]{url}  
\usepackage{graphicx} 
\usepackage{natbib}  
\usepackage{caption} 
\usepackage{algorithm}

\usepackage{algpseudocode}

\usepackage{newfloat}
\usepackage{listings}
\DeclareCaptionStyle{ruled}{labelfont=normalfont,labelsep=colon,strut=off} 
\floatstyle{ruled}
\newfloat{listing}{tb}{lst}{}
\floatname{listing}{Listing}
\usepackage{amsmath}
\usepackage{amssymb}
\usepackage{mathtools}
\usepackage{amsthm}
\usepackage{multirow}

\newcommand{\vecscalar}[1]{\boldsymbol{#1}}
\newcommand{\vecrv}[1]{\mathbf{#1}}
\newcommand{\rv}[1]{\mathrm{#1}}

\newtheorem{theorem}{Theorem}[]
\newtheorem{proposition}[theorem]{Proposition}

\title{Where and How to Attack? A Causality-Inspired Recipe for Generating Counterfactual Adversarial Examples}

\author {
    Ruichu Cai\textsuperscript{\rm 1,2},
    Yuxuan Zhu\textsuperscript{\rm 1},
    Jie Qiao\textsuperscript{\rm 1}\thanks{Corresponding author.},
    Zefeng Liang\textsuperscript{\rm 1},
    Furui Liu\textsuperscript{\rm 3},
    Zhifeng Hao\textsuperscript{\rm 4}
}
\affiliations {
    \textsuperscript{\rm 1}School of Computer Science, Guangdong University of Technology, Guangzhou, China\\
    \textsuperscript{\rm 2}Peng Cheng Laboratory, Shenzhen, China\\
    \textsuperscript{\rm 3}Zhejiang Lab, Hangzhou, China\\
    \textsuperscript{\rm 3}College of Science, Shantou University, Shantou, China\\
    \{cairuichu, iamyuxuanzhu, qiaojie.chn, lzfeng011021\}@gmail.com, liufurui@zhejianglab.com, haozhifeng@stu.edu.cn
}

\begin{document}

\maketitle

\begin{abstract}
Deep neural networks (DNNs) have been demonstrated to be vulnerable to well-crafted \emph{adversarial examples}, which are generated through either well-conceived $\mathcal{L}_p$-norm restricted or unrestricted attacks. Nevertheless, the majority of those approaches assume that adversaries can modify any features as they wish, and neglect the causal generating process of the data, which is unreasonable and unpractical. For instance, a modification in income would inevitably impact features like the debt-to-income ratio within a banking system. By considering the underappreciated causal generating process, first, we pinpoint the source of the vulnerability of DNNs via the lens of causality, then give theoretical results to answer \emph{where to attack}. Second, considering the consequences of the attack interventions on the current state of the examples to generate more realistic adversarial examples, we propose CADE, a framework that can generate \textbf{C}ounterfactual \textbf{AD}versarial \textbf{E}xamples to answer \emph{how to attack}. The empirical results demonstrate CADE's effectiveness, as evidenced by its competitive performance across diverse attack scenarios, including white-box, transfer-based, and random intervention attacks.
\end{abstract}

\section{Introduction}
Deep Neural Networks (DNNs) have achieved tremendous success in various tasks and have been widely used in critical domains such as facial recognition \cite{c:facenet}, medical diagnostics \cite{c:med1}, and autonomous driving \cite{c:autodrive}. Despite their unprecedented achievements, DNNs remain vulnerable to the well-crafted adversarial examples \cite{c:intriging_prop,c:evasion_att}, and there has been a recent thrust on generating adversarial examples through, e.g., $\mathcal{L}_p$-norm restricted attack \cite{c:fgsm,c:bim,c:pgd,c:cw,c:deepfool}, and unrestricted attack \cite{axv:patch,c:sae,c:cadv,c:acgan,c:semanticadv,c:ncf}. 

The $\mathcal{L}_p$-norm approaches reveal DNNs' vulnerability by searching for the perturbation in raw pixel-space within a bounded norm to preserve the photo-realism, while the unrestricted approaches replace such bounded perturbation with, e.g., geometric distortions \cite{c:deftransformation}, color/texture changing \cite{c:sae,c:cadv,c:ncf}, and semantic changing \cite{c:semanticadv, c:acgan}, etc.
Nevertheless, the majority of these methods assume that an attacker can modify any features as they wish, which is unreasonable if we aim to generate an adversarial example in real-world, e.g., the intractability of accessing the digital input to an image recognition model renders those methods perturbing the raw pixel-space fail. 
Moreover, we argue that only altering the alterable features while leaving others unchanged might also be impractical as it ignores the effect caused by the altering features, which has been underappreciated by the majority of the existing methods.

As a motivating example, consider a credit scoring model used by a financial institution to assess the creditworthiness of loan applicants. The model incorporates various features such as income, debt-to-income ratio, and credit history. To produce the adversarial example, it is unreasonable to disturb the income while leaving the debt-to-income ratio unchanged as it is induced by income and debt. This seemingly trivial observation has the underappreciated aspect that a causal generating process should also be involved to produce the adversarial example toward a practical scenario. 

\begin{figure}
    \centering
    \includegraphics[scale=0.339]{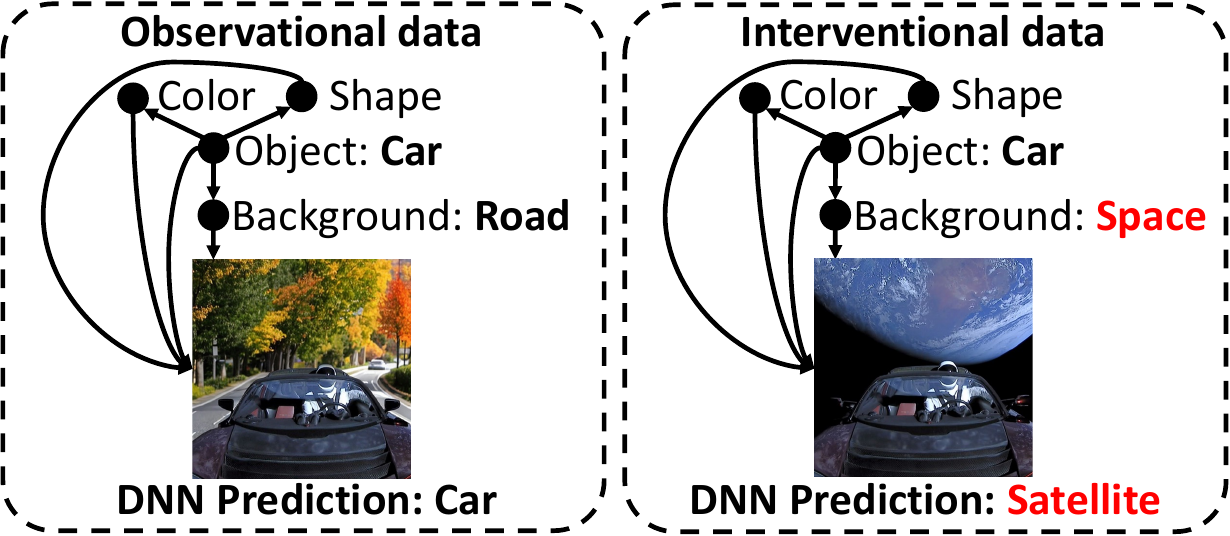}
    \caption{Discriminative DNN's vulnerability to the interventional data.}
    \label{fig:intro}
\end{figure}

In this work, we provide a new perspective view on the adversarial attacks by taking the causal generating process into consideration, and propose a framework, CADE, that can generate \textbf{C}ounterfactual \textbf{AD}versarial \textbf{E}xamples.
We introduce our CADE by answering two fundamental questions: 1) where to attack: understanding the adversarial example from the causal perspective to select valid disturbed variables; 2) how to attack: leveraging the causal generating process to generate more realistic/reasonable adversarial examples, since naively changing the cause variable without changing the effect variables will result in unrealistic examples.
First, to answer where to attack, incorporated with structural information of the data, we give theoretical characterizations of the vulnerability of discriminative DNNs, i.e., the non-robustness to the interventional data, to which human perception is robust thanks to the capability of causal inference. For example, as the data-generating process shown on the left of Figure~\ref{fig:intro}, a car is always on the ground, humans can recognize the car even if it is in space (interventional), while the DNNs recognize it as a satellite since it leverages the background of ``space'' on decision-making.
Addressing this vulnerability, we analyze the effects of interventions explicitly to offer clear guidance for both observable and latent attacks.
Second, to answer how to attack, the key problem is to predict the consequences when variables are intervened given the current observation, and the examples obtained are also known as counterfactuals in the literature of causality. 
For instance, in Figure~\ref{fig:intro}, given the observation (left), when we intervene on background, the consequence is the changed image while other characters of the current observation are preserved (color, shape), where the preserved part is referred to the exogenous representing the current state of the example.
To generate counterfactuals, we resort to the generation framework proposed in \cite{b:causality}, which requires the causal generating process incorporated in.
Thanks to the recent success of causal discovery \cite{c:notears,c:daggnn}, generative modeling \cite{c:vae,axv:gan,c:ddpm}, and causal representation learning \cite{c:causalgan,c:causalvae,a:dear}, it is plausible to recover the generating process and generate counterfactual examples from interventional distribution practically.
By knowing where and how to attack, our CADE offers an executable recipe to generate counterfactual examples.
Empirically, our CADE achieves competitive results on white-box and transfer-based black-box attacks, and non-trivial performance with random intervention where no substitute model is involved.

Overall, our contributions are summarized as follows:
\begin{itemize}
    \item We give a theoretical characterization of the discriminative DNNs' vulnerability via the lens of causality, which offers clear guidance to answer where to attack.
    \item To generate more realistic examples, we propose CADE, a framework that can generate \textbf{C}ounterfactual \textbf{AD}versarial \textbf{E}xamples by considering the consequences of the interventions.
    \item The experimental results prove the effectiveness of our proposed CADE, by achieving competitive results on white-box, transfer-based, and even random attacks.
\end{itemize}

\section{Background}
To reason counterfactual, we adopt the structural causal model (SCM) framework \cite{b:causality} which defines a causal model as a triplet $M(\vecrv{x},f,\vecrv{u})$ over variables $\vecrv{x}=\{\rv{x}_1, \ldots, \rv{x}_d\}$ as: (i) a collection of structural assignments $\left\{\rv{x}_i:=f_i\left(\mathrm{Pa}_i, \rv{u}_i\right)\right\}_{i=1}^d$, where $f_i$ are deterministic functions computing variable $i$ from its causal parents $\mathrm{Pa}_i \subseteq\left\{\rv{x}_1, \ldots, \rv{x}_d\right\} \backslash \rv{x}_i$; and (ii) a factorizing joint distribution over the unobserved noise variables $\vecrv{u}=\{\rv{u}_1, \ldots, \rv{u}_d\}$. Together, (i) and (ii) define a causal generative process and imply an observational joint distribution over $\rv{x}_1, \ldots, \rv{x}_d$ which factorizes over the causal graph $\mathcal{G}$ as:
$
p\left(\rv{x}_1, \ldots, \rv{x}_d\right)=\prod_{i=1}^d p\left(\rv{x}_i \mid \mathrm{Pa}_i\right)
$.
    

\section{Methodology}
In this section, we first explain the existence of adversarial examples from a causal perspective, and give theoretical analyses to answer \emph{where to attack}. Then, we propose CADE, a framework that can generate Counterfactual ADversarial Examples, to answer \emph{how to attack}.

\subsection{Motivating Example}
We start with a motivating example of a linear model to illustrate the existence of adversarial examples via the model coefficients.
We consider a linear data-generating process shown in Eq.~\eqref{eq:toy_gen}.
\begin{equation}
\left\{
\label{eq:toy_gen}
\begin{array}{ll}
      \rv{x}_1=\rv{u}_1 \quad &\rv{u}_1\sim\mathcal{N}(0, \sigma_{1}^2)  \\
      \rv{y} = a\rv{x}_1 + \rv{u}_y \quad &\rv{u}_y\sim\mathcal{N}(0, \sigma_{y}^2) \\
      \rv{x}_2 = b\rv{x}_3 + c\rv{y} + \rv{u}_2 \quad &\rv{u}_2\sim\mathcal{N}(0, \sigma_{2}^2) \\
      \rv{x}_3 = \rv{u}_3 \quad &\rv{u}_3\sim\mathcal{N}(0, \sigma_{3}^2)
\end{array}
,
\right.
\end{equation}
where $\rv{u}_1$, $\rv{u}_y$, $\rv{u}_2$, and $\rv{u}_3$ are the exogenous with zero mean and finite variance.
Considering regressing the target variable $\rv{y}$ with variables $\rv{x}_1$, $\rv{x}_2$ and $\rv{x}_3$ in a linear fashion, that is, $\hat{\rv{y}}=\vecrv{w}^T\vecrv{x}$, where $\vecrv{x}=[\rv{x}_1,\rv{x}_2,\rv{x}_3]^T$ and $\vecrv{w}=[\rv{w}_1, \rv{w}_2,\rv{w}_3]^T$. The model parameter $\vecrv{w}$ is obtained by Empirical Risk Minimization (ERM), which is:
\begin{equation}
\label{eq:solution_w}       
\vecrv{w}=
\left[\begin{matrix}\frac{\sigma_{2}^{2} a}{\sigma_{2}^{2} + \sigma_{y}^{2} c^{2}},& \frac{\sigma_{y}^{2} c}{\sigma_{2}^{2} + \sigma_{y}^{2} c^{2}},& - \frac{\sigma_{y}^{2} b c}{\sigma_{2}^{2} + \sigma_{y}^{2} c^{2}}\end{matrix}\right]^T,
\end{equation}
where the proof of Eq.~\eqref{eq:solution_w} is shown in Appendix~A.1. Since $\rv{u}_2$ has finite variance, we have non-zero $\rv{w}_2$ and $\rv{w}_3$.
However, according to Eq.~\eqref{eq:toy_gen}, the most robust model should be $\vecrv{w}^*=[a,0,0]^T$, i.e., conceptualizing $\rv{y}$ only by $\rv{x}_1$.
We can observe the vulnerability of the model with non-zero $\rv{w}_2$ and $\rv{w}_3$, indicating that we can generate adversarial examples by changing the realization of $\rv{x}_2$ (child of $\rv{y}$), or $\rv{x}_3$ (co-parent of $\rv{y}$). Nevertheless, when interventions are conducted, it is necessary to consider the consequence of each intervention, which the majority of the existing methods ignore. Considering the causal generating process, we analyze the interventional effect of both child ($\rv{x}_2$) and co-parent ($\rv{x}_3$) as follows.

\subsubsection{Children Intervention}
Given an original input example $\vecscalar{x}$, when intervening on the child of $\rv{y}$ (namely, $\rv{x}_2$), i.e., $do(\rv{x}_2=x_2+\eta_2)$, the generated counterfactual adversarial example $\vecscalar{x}^{adv}$ will be $\vecscalar{x}^{adv}=[x_1,x_2+\eta_2,x_3]^T$, which leads to an adversarial output:
\begin{equation*}
\vecrv{w}^T\vecscalar{x}^{adv}=\vecrv{w}^T\vecscalar{x}+\rv{w}_2\eta_2,
\end{equation*}
where $\vecrv{w}$ is shown in Eq.~\eqref{eq:solution_w}. The intervention on the child variable ($\rv{x}_2$) causes the shifted adversarial output by $\rv{w}_2\eta_2$.

\subsubsection{Co-parents Intervention}
Regarding the intervention on \emph{co-parents} of $\rv{y}$ (namly, $\rv{x}_3$), that is, $do(\rv{x}_3=x_3+\eta_3)$, the counterfactual adversarial will be $\vecscalar{x}^{adv}=[x_1,x_2+b\eta_3,x_3+\eta_3]^T$, rendering the adversarial output to be:
\begin{equation*}
\vecrv{w}^T\vecscalar{x}^{adv}=\vecrv{w}^T\vecscalar{x}+\rv{w}_2 b\eta_3+\rv{w}_3\eta_3=\vecrv{w}^T\vecscalar{x}.
\end{equation*}

The intervention on the co-parent variables ($\rv{x}_3$) does not cause any damage to the output, where the shifted term $w_3\eta_3$ is canceled out by the interventional effect of $\rv{x}_3$ to its child variable $\rv{x}_2$, which is $\rv{w}_2 b\eta_3$. 
Intuitively, intervening co-parent variables ($\rv{x}_3$) does not change any structural mechanism as in Eq.~\eqref{eq:toy_gen}, which the linear model has already modeled, thus causing no shift of the output.

\subsection{Generating Adversarial Example: Where to Attack?}
The linear motivating example suggests an intuition of the vulnerability of the discriminative model via the coefficients. 
Despite the linear model, the theoretical analysis of this vulnerability of non-linear model (e.g., DNNs) is further investigated from a probabilistic view in this section.
Specifically, we first give our analysis on observable variables, then extend the results to the latent variables in which the causal variables are not observable (e.g. the objects in the image).

\subsubsection{Observable Variable Intervention}
From a probabilistic view, a discriminative DNN aims to approximate the conditional distribution $p_{M}(\rv{y}|\vecrv{x})$ via ERM, where $M$ denotes the SCM that parameterizes the generating process of $\vecrv{x}$ and $\rv{y}$.
With the great capacity, DNN can well approximate a distribution $p_{\theta}(\rv{y}|\vecrv{x})\approx p_M(\rv{y}|\vecrv{x})$ where $\theta$ denotes the parameter of DNN, but exhibits limitation in generalizing to shifted distribution \cite{axv:irm,c:tiva}.
Here we analyze how $\rv{y}$ is predicted given the observable $\vecrv{x}$ and how to obtain a shifted distribution by intervention to fool the model.
First, given the observable $\vecrv{x}$ to predict the target $\rv{y}$, the conditional distribution can be derived as the following Proposition~\ref{prop:yx_to_ymb}.

\begin{proposition}
\label{prop:yx_to_ymb}
Given the SCM $M$, the discriminative conditional distribution:
\begin{equation*}
\label{eq:y_given_mb}
p_{M}(\rv{y}|\vecrv{x})=p_M(\rv{y}|\mathrm{Mb}_y^{M}).
\end{equation*}
where $\mathrm{Mb}_y^{M}$ denotes the Markov blanket of $\rv{y}$ under SCM $M$, including the parents, children, co-parents of $\rv{y}$.
\end{proposition}
The proof of Proposition~\ref{prop:yx_to_ymb} is given in Appendix~A.2.
Proposition~\ref{prop:yx_to_ymb} suggests that the $\mathrm{Mb}_y^{M}$ are only variables needed to predict $\rv{y}$ given the observable $\vecrv{x}$.
However, in the literature of causal inference, the most robust way of conceptualizing the target variable $\rv{y}$ is to only use its parent variables. 
Thus, the dependencies between $\rv{y}$ and $\mathrm{Mb}_y^{M}$ reveal the vulnerability of discriminative DNNs, offering adversaries an opportunity to attack by leveraging this property.
When generating adversarial examples, intervening on $\rv{y}$ and its parents do not correspond to the attack since it changes the $\rv{y}$, thus the remaining children and co-parents of $\rv{y}$ are only valid variables we can control. To answer \emph{where to attack}, we theoretically analyze the effect of the intervention on children and co-parents separately as follows.

\begin{proposition}
\label{prop:intervention_observe}
Given a SCM $M(\{\vecrv{x},\rv{y}\},f,\vecrv{u})$ where the underlying conditional distribution is $p_{M}(\rv{y}|\vecrv{x})$, if an intervention is conducted on a $\rv{y}$'s child $\rv{x}_j$ resulting in a new SCM $M'(\{\vecrv{x},\rv{y}\},f',\vecrv{u})$ with $f'=\{f_i| i\ne j\}\cup\{f'_j\}$, then we get nonequivalent interventional distribution $p_{M'}(\rv{y}|\vecrv{x})\neq p_{M}(\rv{y}|\vecrv{x})$; and if an intervention is conducted on a $\rv{y}$'s co-parent but not child variable $\rv{x}_k$ resulting in a new SCM $M'(\{\vecrv{x},\rv{y}\},f',\vecrv{u})$ with $f'=\{f_i: | i\ne k\}\cup\{f'_k\}$, then the interventional distribution $p_{M'}(\rv{y}|\vecrv{x})=p_{M}(\rv{y}|\vecrv{x})$.
\end{proposition}

The proof of Proposition~\ref{prop:intervention_observe} is given in Appendix~A.3.
Proposition~\ref{prop:intervention_observe} states the nonequivalence in distributions between $p_{M'}(\rv{y}|\vecrv{x})$ and $p_{M}(\rv{y}|\vecrv{x})$ when \emph{children} are intervened, and the equivalence between $p_{M'}(\rv{y}|\vecrv{x})$ and $p_{M}(\rv{y}|\vecrv{x})$ when \emph{co-parents but not children} are intervened.
Intuitively, the underlying rationale for the inequality lies in the disruption caused to the internal structure of $\mathrm{Mb}_{y}^{M}$.
Since the distribution $p_{\theta}(\rv{y}|\vecrv{x})$ learned by DNN is to approximate $p_{M}(\rv{y}|\vecrv{x})$ under the SCM $M$, it exhibits limitations in generalizing to the shifted interventional distribution $p_{M'}(\rv{y}|\vecrv{x})$. 
This incapability of generalization offers a clear answer to where to attack, i.e., crafting adversarial examples drawn from a shifted interventional distribution with $\rv{y}$ preserved. 
This can be done by, such as children intervention, or both children and co-parents interventions that can damage the inner mechanisms within $\mathrm{Mb}_{y}^{M}$, suggested by Proposition~\ref{prop:intervention_observe}.

\subsubsection{Latent Variable Intervention}
When facing images, most of the existing methods modify them in the raw pixel space, which is, however, impractical and highly-cost in the real-world. 
To mitigate this issue and keep the realism of the generated examples, it is plausible to attack the latent variables $\vecrv{z}$ with semantics that determines $\vecrv{x}$, where each variable of $\vecrv{z}$ can be causal-related, and the target $\rv{y}$ is included in $\vecrv{z}$ that determines $\vecrv{x}$ \cite{c:tcas}. 
Since the image $\vecrv{x}$ is a child of $\vecrv{z}$, each variable of $\vecrv{z}$ except $\rv{y}$ becomes co-parent of $\rv{y}$, indicating that when an image $\vecrv{x}$ is given to predict $\rv{y}$, $\rv{y}$ is correlated with every other variable of $\vecrv{z}$. 
To answer \emph{where to attack}, we investigate the connection between $\vecrv{z}$ and the conditional distribution $p_{M}(\rv{y}|\vecrv{x})$ through the following proposition.

\begin{proposition}
\label{prop:latent}
Given a SCM $M(\{\vecrv{x},\vecrv{z}\}, \allowbreak \{f,g\}, \{\vecrv{u}_x,\vecrv{u}_z\})$ where $\rv{z}_i=f_i(\mathrm{Pa}_i^M, \rv{u}_i)$, $\vecrv{x}=g(\vecrv{z},\vecrv{u}_x)$, and target $\rv{y}$ is included in $\vecrv{z}$, when interventions are conducted on the latent $\vecrv{z}$ to obtain a new SCM $M'(\{\vecrv{x},\vecrv{z}\},\{f',g\},\{\vecrv{u}_x,\vecrv{u}_z\})$, if the conditional distribution $p_{M'}(\rv{y}|\vecrv{x})\neq  p_{M}(\rv{y}|\vecrv{x})$, then $p_{f'}(\vecrv{z})\neq p_{f}(\vecrv{z})$.
\end{proposition}

Though $p_{f'}(\vecrv{z})\neq p_{f}(\vecrv{z})$ is a necessary condition for $p_{M'}(\rv{y}|\vecrv{x})\neq  p_{M}(\rv{y}|\vecrv{x})$ according to Proposition~~\ref{prop:latent}, we analyze that at most cases, $p_{f'}(\vecrv{z})\neq p_{f}(\vecrv{z})$ can yield $p_{M'}(\rv{y}|\vecrv{x})\neq  p_{M}(\rv{y}|\vecrv{x})$, and the detailed proof and analysis of Proposition~\ref{prop:latent} is given in Appendix~A.4. 
Since the DNN only fits the $p_{M}(\rv{y}|\vecrv{x})$, we can generate the adversarial example drawn from a shifted $p_{f'}(\vecrv{z})$.
Since the joint can be factorized as $p_{f}(\vecrv{z})=\prod_{i=1}^n p\left(\rv{z}_i \mid \mathrm{Pa}_i\right)$, a new mechanism $f'$ obtained by interventions that cause the structural change of $\vecrv{z}$ can result in a shifted $p_{f'}(\vecrv{z})$.
To preserve $\rv{y}$, interventions on $\rv{y}$ and its parents do not correspond to the attack we consider, and some possible choices can be variables that changed the structure of $\vecrv{z}$ except for those two, such as the children of $\rv{y}$.

\subsection{Generating Adversarial Example: How to Attack?}
Knowing where to attack, the next step is to generate the adversarial example by considering the consequence of each intervention on the current state, since intervening on one variable will inevitably cause its descendants to change.
The generated example, also called counterfactual, is the consequences under a hypothetical scenario where interventions are conducted, given the original example. To generate it, we resort to the framework proposed in \cite{b:causality,b:why}, which requires three fundamental steps: 1) abduction, 2) action, and 3) prediction.
Here, we introduce the process for observable $\vecrv{x}$, and adapt it for latent $\vecrv{z}$ in a similar fashion.
First, to properly parameterize the causal generating process of an SCM $M$ and compute the interventional effect efficiently, we can adopt the general non-linear generating process proposed in \cite{c:daggnn}, which is:
\begin{equation}
\label{eq:nlr_scm_x}
f(\vecrv{x}) = \vecrv{A}^Tf(\vecrv{x})+\vecrv{u},
\end{equation}
where $f$ denotes an invertible non-linear function, and $\vecrv{A}$ denotes the weighted adjacency matrix of the causal DAG.

\subsubsection{Abduction}
The abduction step aims to maintain the characters of the current state of the given observation, by recovering the exogenous $\vecrv{u}$. From Eq.~\eqref{eq:nlr_scm_x}, the exogenous $\vecrv{u}$ can be recovered by:
\begin{equation}
\label{eq:abduction}
\vecrv{u} = (\vecrv{I}-\vecrv{A}^T)f(\vecrv{x}).
\end{equation}

\subsubsection{Action (Intervention) and Prediction} 
Formally, these two processes are as follows. First, intervene on the desired variables $\vecrv{x}_S$ to obtain $\vecrv{x}'$, where $\vecrv{x}_S$ is obtained from the variable selection process depicted in Figure~\ref{fig:framework} and $\vecrv{x}'$ denotes the full input vector with $\vecrv{x}_S$ changed.
Then, predict the consequence of each intervention to obtain the corresponding adversarial example $\vecrv{x}^{adv}$.
One way to incorporate these two processes is through the inversion version of Eq.~\eqref{eq:nlr_scm_x}, i.e., $\vecrv{x}^{adv} = f^{-1}((\vecrv{I}-\vecrv{A}^T)^{-1}\vecrv{u}')$, where $\vecrv{x}^{adv}$ is obtained by $\vecrv{u}'$ with $\vecrv{u}_S$ intervened.
However, the matrix inversion process of $(\vecrv{I}-\vecrv{A}^T)^{-1}$ will introduce extra error and computational cost. To remedy this issue, we directly intervene on $\vecrv{x}_S$ to obtain $\vecrv{x}^{adv}$ with the help of the mask $\vecrv{m}$:
\begin{equation}
\label{eq:act_pre}
\begin{aligned}
\vecrv{x}^{adv}=&f^{-1}(\vecrv{A}^Tf(\vecrv{x}')\odot (1-\vecrv{m})\\
&+f(\vecrv{x}')\odot \vecrv{m}+ \vecrv{u}\odot (1-\vecrv{m})),
\end{aligned}
\end{equation}
where $\vecrv{m}$ is a binary mask with $\vecrv{m}_S=1$ indicating variable $\vecrv{x}_S$ is intervened. 
Intuitively, the first term in the first bracket of r.h.s. of Eq.~\eqref{eq:act_pre} aims to update the effect of each intervention and set the realization of the intervened variables to 0. The second term aims to add the realization of the intervened variables back. The third term aims to set the exogenous of the intervened variables to 0.

\subsubsection{White-Box Attack}
Under a white-box setting, we can leverage the gradient information of the target model to guide the interventions. Attacking a discriminative DNN, from the probabilistic perspective, our goal is to generate an adversarial example $\vecrv{x}^{adv}$ that can shift the conditional probability $p(\rv{y}|\vecrv{x}^{adv})$, which can be accomplished by maximizing the prediction loss:
\begin{equation}
\label{eq:obj_white}
\mathop{\max}_{\vecrv{x}^{adv}\in\mathcal{X}'}{\mathcal{L}_{pred}(f_{\theta}(\vecrv{x}^{adv}), \rv{y})},
\end{equation}
where $\mathcal{X}'$ denotes the sample space under a specific interventional distribution, and $f_{\theta}(.)$ denotes the output of DNN. To search on $\mathcal{X}'$ to obtain $\vecrv{x}^{adv}$, we only update the selected variables $\vecrv{x}_S$ by freezing the gradients of others except for $\vecrv{x}_S$ in the \emph{action} step, then update the consequence of each intervention according to \eqref{eq:act_pre} in the \emph{prediction} step, as depicted in Figure~\ref{fig:framework}. The algorithm is given in Appendix~B.3.

\subsubsection{Black-Box Attack}
The obtained white-box adversarial examples then can be transferred to other victim models thanks to the transferability property \cite{c:intriging_prop}.
Further, our CADE can be applied to a more strict scenario where the substitute model is absent.
Benefit from the causal information, it is plausible to generate query-free adversarial examples without any white-box gradient information, by addressing the variables that can shift the conditional probability $p(\rv{y}|\vecrv{x})$, according to Propositions~\ref{prop:intervention_observe}, and \ref{prop:latent}.
To achieve this, one way is to add random noises to those effective variables or reassign random realizations to them.

\subsubsection{Latent Attack for Image}
For image data, we assume the images are generated according to a two-level "latent variable-image" model and we attack on the latent variables. When attacking, we first recover the causally related latent variables $\vecrv{z}=Encoder(\vecrv{x})$ from the images using an encoder step. Then, following \eqref{eq:abduction} and \eqref{eq:act_pre}, we obtain the adversarial latent code $\vecrv{z}^{adv}$ by replacing the observable $\vecrv{x}$ with the latent $\vecrv{z}$. We finally obtain the adversarial example through $\vecrv{x}^{adv}=Decoder(\vecrv{z}^{adv})$. The detailed implementation and algorithm are given in Appendix~B.2 and B.3.

\begin{figure}[t]
    \centering
    \includegraphics[scale=0.44]{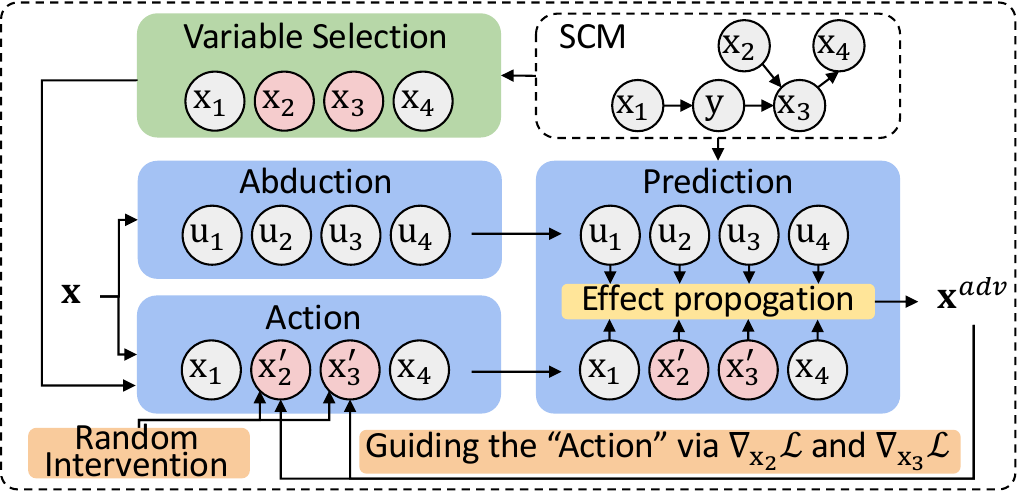}
    \caption{Framework of CADE.}
    \label{fig:framework}
\end{figure}

\section{Experiment}

\subsection{Experimental Setup}

\subsubsection{Dataset}
\begin{figure}[h]
    \centering
    \includegraphics[scale=0.565]{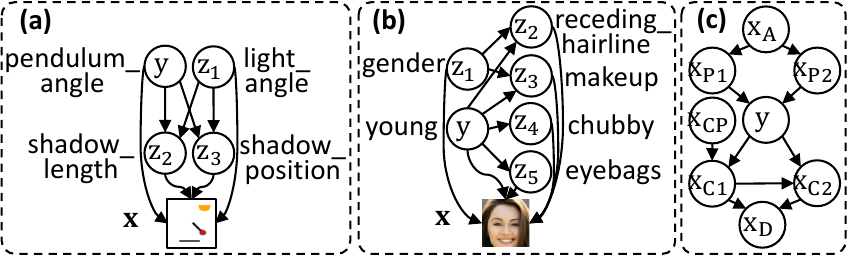}
    \caption{The causal graphs for each dataset: (a) \emph{Pendulum}, (b) \emph{CelebA(Attractive)}, and (c) \emph{SynMeasurement}.}
    \label{fig:dgp}
\end{figure}
We evaluate our approach on three datasets, \emph{Pendulum} \cite{c:causalvae}, \emph{CelebA} \cite{c:celeba}, and a synthetic measurement dataset, denoted as \emph{SynMeasurement}. 
\emph{Pendulum}, a synthetic image dataset generated by four causally related continuous variables, which follow the physical mechanism depicted in Figure~\ref{fig:dgp}~(a). 
\emph{CelebA}, a real-world human face dataset with 40 labeled binary attributes has been well investigated in generative modeling \cite{c:stargan,c:starganv2}, including causal generative modeling \cite{c:causalgan,c:causalvae,a:dear}. In our experiments, we deploy our CADE using a causal graph proposed in \cite{a:dear}, which is depicted in Figure~\ref{fig:dgp}~(b). 
To further validate our theoretical characterization of CADE, we introduce a synthetic measurement dataset, \emph{SynMeasurement}, whose corresponding causal graph is depicted in Figure~\ref{fig:dgp}~(c). More details about the generating process of those are given in Appendix~B.1. 

\subsubsection{Setup for Pendulum and CelebA}
We select Res-50 \cite{c:resnet}, VGG-16 \cite{c:vgg}, and their respective defense variants (adversarial trained with PGD \cite{c:pgd}, shortly denoted as Res-50(D) and VGG-16(D)) as our classification model. For \emph{Pendulum}, our task is to predict the \emph{pendulum\_angle} where the discrete categorical target labels are converted from the continuous \emph{pendulum\_angle} based on to which angle intervals an image belongs. 
To make the dataset more realistic, we introduce random noises on \emph{pendulum\_angle} on 15~\% of data when generating images, representing the measurement error.
For \emph{CelebA}, our task is to predict if a person is young or not.
To recover the causal latent representations depicted in Figure~\ref{fig:dgp}~(a),~(b), we leverage a state-of-the-art causal generative model \cite{a:dear}, and interventions on such representation are performed to generate counterfactual adversarial examples.

We evaluate the effectiveness of the attacks by reporting the attack success rates (ASR) of white-box, transfer-based, and random black-box attacks. 
For comparison, we select various state-of-the-art attack methods including PGD \cite{c:pgd}, C\&W \cite{c:cw}, SAE \cite{c:sae}, ACE \cite{c:ace}, APGD($\mathcal{L}_{\inf}$) \cite{c:apgd}, and APGD($\mathcal{L}_{1}$) \cite{c:apgdl1} as our baselines.
For \emph{Pendulum}, we compare the results of our CADE intervening on different variables, i.e., CADE(1) on \emph{light\_angle} (co-parent of target), CADE(2) on \emph{shadow\_length} (child of target), CADE(3) on \emph{shadow\_position} (child of target), and CADE(Mb) (Markov blanket of target).
For \emph{CelebA}, we implement our CADE by intervening on $\vecrv{z}_{1:5}$ depicted in Figure~\ref{fig:dgp}~(b).
Details of such implementation are shown in Appendix~B.

\subsubsection{Setup for SynMeasurement}
We select Linear, MLP, and their respective adversarial trained (PGD) defense variants, denoted as Linear(D) and MLP(D) as our regression model to predict $\rv{y}$.
We test our CADE in a finer-grained setting where the CP (co-parent), C1 (child 1), and C1+C2 (child 1+2) interventions are conducted separately. Further, we investigate our CADE w/ and w/o the prediction process in counterfactual generation. Specifically, we use the term ``intervention'' (denoted as (i) shortly) to refer to the attacks w/ the prediction process, and ``perturbation'' (denoted as (p)) to refer to the attacks w/o the prediction process. 

\subsection{Attacks on Pendulum}

\subsubsection{Quantitative Analysis}
We compare our proposed CADE with various state-of-the-art baselines, where the results of attack success rate are reported in Table~\ref{tab:asr_pend}.

Regarding the results compared with the baselines, we can observe that the adversarial examples obtained by CADE(Mb) achieve the highest transfer-based ASR among all competitors. Specifically, our CADE(Mb), CADE(2), CADE(3) not only achieve high ASR on the standard trained model but also achieve consistently high scores on the adversarial trained defense model, which indicates the effectiveness of our resulting adversarial examples to reveal the vulnerability of both standard and adversarial trained models. Besides, our CADE(Mb) also achieves competitive ASR on the white-box scenario, which is 99.4 and 99.7 under Res-50 and VGG-16, respectively.
Further, our CADE with random intervention and no substitute model achieves competitive results compared to baselines, and even becomes the first or second winner on the transfer-based results.

Regarding the results of our CADE intervening on different variables, i.e., CADE(1), CADE(2), CADE(3), and CADE(Mb), first, we observe that the ASR of CADE(1) which intervenes on co-parent of $\rv{y}$ is lower than those of CADE(2) and CADE(3) which intervenes on the child of $\rv{y}$.
The lowest result of CADE(1) can be suggested by Proposition~\ref{prop:latent} with no structural change in $\vecrv{z}$. 
However, the vulnerability of models to the CADE(1) can be due to, the white-box gradient exploiting the weakness of the model, and the models that overfit to the perfectly generated data tend to make mistakes when the imperfect legitimate examples are generated from the generative model with noises, which is also suggested in \cite{c:acgan}.
Further, we provide the result of the attack using \emph{Pendulum} simulator in Appendix~C.2.

\begin{table}[t]
    \centering
    \scalebox{0.905}{
\begin{tabular}{cccccc}
\hline
               & Attacks                    & R50          & R50(D)     & V16          & V16(D)     \\ \hline
\multirow{10}{*}{\rotatebox[origin=c]{90}{Res-50}} 
                         & PGD                        & \textbf{100.0*} & 0.3           & 0.8             & 0.2           \\
                         & C\&W                       & \textbf{100.0*} & 0.3           & 0.0             & 0.0           \\
                         & SAE                        & \textbf{100.0*} & 37.2          & 83.5            & 46.0          \\
                         & ACE                        & \textbf{100.0*} & 0.3           & 22.1            & 1.0           \\
                         & APGD($\mathcal{L}_{\inf}$) & \textbf{100.0*} & 0.3           & 0.3             & 0.2           \\
                         & APGD($\mathcal{L}_{1}$)    & \textbf{100.0*} & 57.2          & 55.8            & 47.3          \\ \cline{2-6} 
                         & CADE(1)                    & 75.3*           & 72.8          & 76.8            & 75.6          \\
                         & CADE(2)                    & 98.6*           & 96.3          & \textbf{98.9}   & 98.9          \\
                         & CADE(3)                    & 95.0*           & 93.9          & 94.7            & 92.6          \\
                         & CADE(Mb)                   & 99.4*           & \textbf{98.6} & \textbf{98.9}   & \textbf{99.2} \\ \hline
\multirow{10}{*}{\rotatebox[origin=c]{90}{VGG-16}} 
                         & PGD                        & 5.8             & 0.5           & \textbf{100.0*} & 0.0           \\
                         & C\&W                       & 2.9             & 0.3           & 99.8*           & 0.0           \\
                         & SAE                        & 24.4            & 1.8          & \textbf{100.0*} & 10.4          \\
                         & ACE                        & 26.4            & 2.1           & \textbf{100.0*} & 1.3           \\
                         & APGD($\mathcal{L}_{\inf}$) & 5.4             & 0.5           & \textbf{100.0*} & 0.0           \\
                         & APGD($\mathcal{L}_{1}$)    & 78.7            & 27.9          & \textbf{100.0*} & 63.9          \\ \cline{2-6} 
                         & CADE(1)                    & 70.7            & 74.2          & 75.5*           & 72.6          \\
                         & CADE(2)                    & 97.4            & 97.4          & 97.9*           & 98.2          \\
                         & CADE(3)                    & 93.1            & 95.2          & 95.2*           & 92.0          \\
                         & CADE(Mb)                   & \textbf{98.9}   & \textbf{99.7} & 99.7*           & \textbf{99.4} \\ \hline
\multirow{4}{*}{\rotatebox[origin=c]{90}{$\times$}}    & CADE(1)                    & 57.7        & 59.0      & 58.2        & 58.9      \\
                         & CADE(2)                    & 84.2        & 83.2      & 81.9        & 82.4      \\
                         & CADE(3)                    & 72.2        & 74.6      & 72.1        & 70.2      \\
                         & CADE(Mb)                   & 87.6        & 87.3      & 86.8        & 86.6      \\ \hline
\end{tabular}
}
    \caption{ASR (\%) on Pendulum. * denotes white-box results, $\times$ denotes results without substitute models. R50 and V16 refer to Res-50 and VGG-16, respectively.}
    \label{tab:asr_pend}
\end{table}

\begin{figure*}[t]
    \centering
    \includegraphics[scale=0.34]{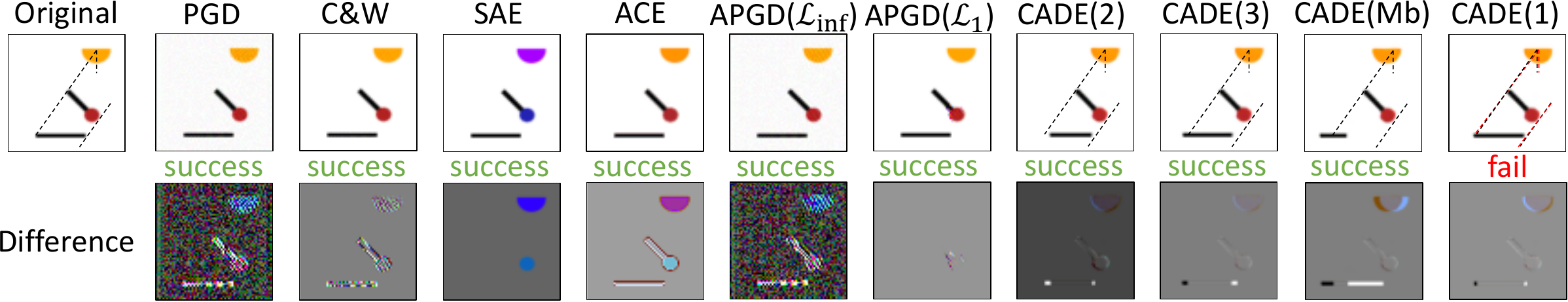}
    \caption{Visualization of adversarial examples against Res-50 on Pendulum obtained by different approaches. The black dash-line highlights the original projection trajectory, while the red dash-line highlights the intervened projection trajectory.}
    \label{fig:qualitative_pend}
\end{figure*}

\subsubsection{Qualitative Analysis}
The adversarial examples obtained by different approaches are shown in Figure~\ref{fig:qualitative_pend}. 
Regarding the examples obtained by our CADE, we make the following observations. 
First, the example obtained by CADE(2) with shadow length shorter, the example obtained by CADE(3) with shadow position shifted, and the example obtained by CADE(Mb) with light angle, shadow length, shadow position changed can successfully fool the DNNs.
Second, the example obtained by CADE(1) with light angle intervened, has caused the shadow changed based on the generative mechanism, rendering it fail to fool the DNNs.

\subsection{Attacks on CelebA}

\subsubsection{Quantitative Analysis}

\begin{table}[t]
    \centering
    \scalebox{0.905}{
\begin{tabular}{cccccc}
\hline
              & Attacks & R50          & R50(D)     & V16          & V16(D)     \\ \hline
\multirow{7}{*}{\rotatebox[origin=c]{90}{Res-50}} 
                        & PGD     & \textbf{100.0*} & 0.5           & 51.2            & 0.2           \\
                        & C\&W    & \textbf{100.0*} & 0.5           & 0.8             & 0.2           \\
                        & SAE     & 77.0*           & 24.1          & 16.9            & 19.5           \\
                        & ACE     & 99.8*           & 21.7          & 15.3            & 18.3          \\
                        & APGD($\mathcal{L}_{\inf}$) & \textbf{100.0*} & 0.6 & 49.7            & 0.2 \\
                        & APGD($\mathcal{L}_{1}$)    & 92.5*           & 0.5 & 14.4             & 0.2 \\
                        & CADE    & 75.8*           & \textbf{47.0} & \textbf{52.7}   & \textbf{47.2} \\ \hline
\multirow{7}{*}{\rotatebox[origin=c]{90}{VGG-16}}
                        & PGD     & 41.1            & 0.6           & {97.3*} & 0.2           \\
                        & C\&W    & 0.9             & 0.5           & \textbf{100.0*} & 0.2           \\
                        & SAE     & 23.4            & 23.7          & 90.2*           & 23.6          \\
                        & ACE     & 18.8            & 16.7          & \textbf{100.0*} & 14.4          \\
                        & APGD($\mathcal{L}_{\inf}$) & 33.9            & 0.5 & \textbf{100.0*} & 0.2 \\
                        & APGD($\mathcal{L}_{1}$)    & 3.0             & 0.5 & 99.8*           & 0.2 \\
                        & CADE    & \textbf{47.0}   & \textbf{41.6} & 80.6*           & \textbf{44.7} \\ \hline
\rotatebox[origin=c]{90}{$\times$}                    & CADE    & 25.9        & 25.8      & 26.3        & 25.0      \\ \hline
\end{tabular}
}
    \caption{ASR (\%) on CelebA. * denotes white-box results, $\times$ denotes results without substitute models. R50 and V16 refer to Res-50 and VGG-16, respectively.}
    \label{tab:quantitative_celeba}
\end{table}
We compare our CADE with several baselines where the ASR are reported in Table~\ref{tab:quantitative_celeba}.
First, we observe that our CADE guided by the causal graph in Figure~\ref{fig:dgp}~(b) outperforms the baselines under the transfer-based black-box setting on both standard and defense models. Second, we observe our CADE does not achieve the best on white-box, this can be due to, the causal graph provided in Figure~\ref{fig:dgp}~(b) is incomplete, and examples drawn from distribution obtained from intervening on such incomplete SCM have limited capacity to flip the prediction. Further, the random interventions with no substitute achieve the highest results on the defense model compared with baselines. 

\subsubsection{Qualitative Analysis}
\begin{figure*}
    \centering
    \resizebox{0.95\linewidth}{!}{
    \includegraphics{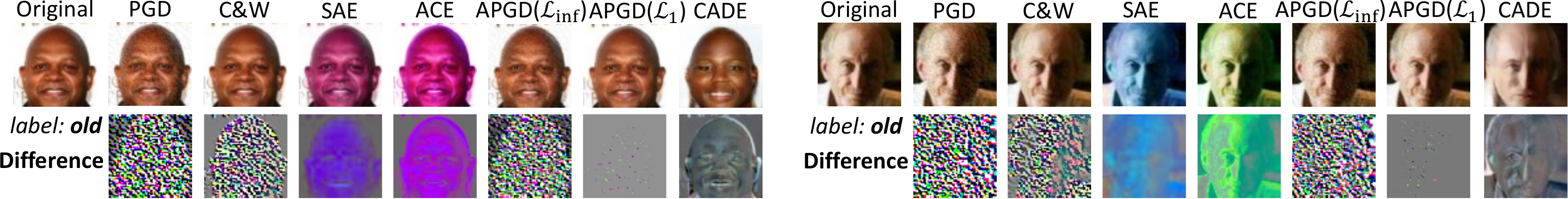}
    }
    \caption{Visualization of adversarial examples against Res-50 on CelebA obtained by different approaches.}
    \label{fig:qualitative_celeba}
\end{figure*}

Figure~\ref{fig:qualitative_celeba} shows the generated adversarial examples against Res-50 of different methods.
We observe that compared with baselines, the examples generated by our CADE have reasonably-looking appearances with latent semantics intervened instead of naively adding noises in the pixel-space. 
Also, we can get some intuitions of why the attack success from the visualizations, e.g., on the left of Figure~\ref{fig:qualitative_celeba}, examples generated by CADE look more feminist (gender) but still bald (receding hairline), suggesting that it could be drawn from an interventional distribution (that females can be bald) to which the DNNs cannot generalize.
Further, we showcase additional case study in Appendix~C.

\subsection{Attacks on SynMeasurement}
\begin{figure*}[t]
    \centering
    \includegraphics[scale=0.67]{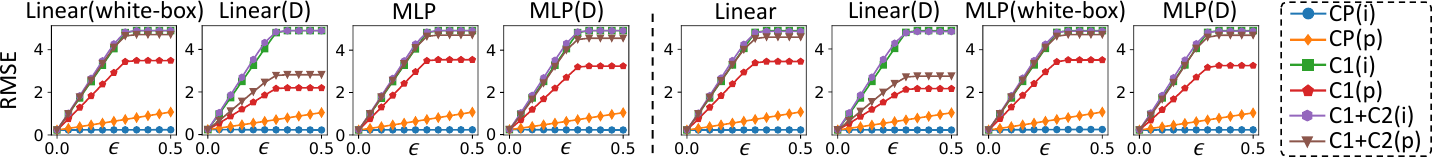}
    \caption{RMSE w.r.t. budget $\epsilon$ of various interventions/perturbations.}
    \label{fig:quantitative_syn}
\end{figure*}

We evaluate our CADE in a finer-grained setting where the CP (co-parent), C1 (child 1), and C1+C2 (child 1+2) interventions are conducted separately, and further compare the results of both intervention (w/ prediction process) and perturbation (w/o prediction process), where the perturbation can be seen as a degradation to the majority of the existing methods, i.e, naively adding noises, and the RMSE results are reported in Figure~\ref{fig:quantitative_syn}.

\subsubsection{Child and Co-parent Interventions}
Regarding the intervention, CP(i), C1(i), C1+C2(i), from Figure~\ref{fig:quantitative_syn} we can make the following observations.
First, the intervention on co-parent, CP(i), makes no effect on attacking the threat models under both black-box and white-box scenarios, which can be well suggested by Proposition~\ref{prop:intervention_observe}. 
Second, both child interventions, C1(i) and C1+C2(i), achieve remarkable results with high RMSE scores. More specifically, intervention on one child C1(i) achieves a competitive result with the intervention on two children, C1+C2(i) on both standardly trained and adversarially trained models, which suggests a way of efficient attacks with lesser costs.

\subsubsection{Intervention and Perturbation}
Comparing the intervention and perturbation results shown in Figure~\ref{fig:quantitative_syn}, we observe an interesting result where the interventions, C1(i) and C1+C2(i), achieve higher RMSE than their respective perturbations, C1(p), and C1+C2(p), suggesting that it is possible to achieve better attacks with lesser costs. This can be due to, the unchanged consequences of the perturbation, which contribute negative effects to the attacks.

\section{Related Work}

\paragraph{Adversarial Robustness}
Here we briefly summarize the existing adversarial attack and defense approaches. For adversarial attack, one line of research lies on adding imperceptible noises within a bounded norm, for example, \cite{c:fgsm} searches for the perturbation direction by one-step gradient, \cite{c:deepfool} produces closer adversarial example by $\mathcal{L}_2$ minimum perturbation, and \cite{c:pgd} aims to find ``most-adversarial'' example by search for the local maximum loss value.
Another line of works goes beyond the bounded perturbation, replacing it with, such as, \cite{c:acccrime} fools a face-recognition model by wearing well-crafted eyeglasses frames, 
\cite{c:naturalexample} search for examples in the vicinity of latent space, and similarly \cite{c:acgan} generate class-conditioned examples in latent space.
As for adversarial defense, one line of works focuses on defense at the training stage, including data augmentation, and adversarial training \cite{c:pgd,c:attndef,c:feat_denoising,c:notbugsfeat}.
Also, another line of works relies on input pre-processing, including JPEG compression, cropping, rescaling, etc \cite{c:deftransformation,axv:jpeg,c:featsqueezing}.
Further, a new line of research focuses on causality-inspired defense, e.g., \cite{c:cama} improve the robustness of DNNs by test-time fine-tuning on unseen perturbation, which is explicitly modeled from a causal view, and \cite{c:causaladv} improve the robustness by penalizing the DNNs incorporated with the restricted attack, where such perturbation is modeled from a causal perspective. However, they do not propose a proper way to generate adversarial examples, which is one of the main differences between ours and those methods.

\paragraph{Causal Learning}
Causal inference has a long history in statistics \cite{b:causality}, and there is now increasing interest in solving crucial problems of machine learning that benefit from causality \cite{avx:causalsurvey}. For instance, causal discovery methods \cite{c:notears,c:daggnn,cai2018self,ijcai2023p0633} identify the underlying causal structure from observations, causal representation learning \cite{c:causalgan,c:causalvae,a:dear} connect causal factors to high-dimensional observations, and several works show the benefits of causality in various applications, including domain adaptation \cite{c:tcas,c:invrep,cai2019learning}, adversarial robustness, and reinforcement learning \cite{c:bandconf,c:cfpolicy}, etc.

\section{Conclusion and Discussion}
In this work, we propose CADE, a causality-inspired framework that can generate \textbf{C}ounterfactual \textbf{AD}versarial \textbf{E}xamples, by considering the underappreciated causal generating process. 
First, we reveal the vulnerability of discriminative DNNs to examples drawn from a nonequivalent interventional distribution with the equivalent target $\rv{y}$, then we provide an explicit recipe to answer where to attack: intervening on variables that renders distribution shift while preserving the consistent target $\rv{y}$.
Second, to generate more realistic examples, we consider the consequence of each intervention on the current state, then propose CADE to generate counterfactual adversarial examples, answering how to attack.
Our experiments demonstrate the effectiveness of CADE, by achieving competitive results on white-box and transfer-based attacks, even non-trivial results on random intervention where the substitute model is not required.

Our CADE is effective, however, obtaining the required causal generating process remains an open challenge.
Our framework should be adapted to more realistic settings where the full generating process can not be obtained. 
In future work, we will explore such settings with limited causal knowledge, in which only partial causal knowledge can be obtained from causal discovery algorithms or expert knowledge. Further, we will verify our framework beyond the digital simulation, and adapt it to the physical world. 
Overall, we believe this work would open up new research opportunities and challenges in the field of adversarial learning, potentially inspiring new designs of defense mechanisms.


\section*{Acknowledgments}
This research was supported in part by National Key R\&D Program of China (2021ZD0111501), National Science Fund for Excellent Young Scholars (62122022), Natural Science Foundation of China (61876043, 61976052), the major key project of PCL (PCL2021A12). 

\bibliography{aaai24}

\onecolumn
\appendix
\setcounter{secnumdepth}{2}
\section*{Appendix of ``Where and How to Attack? A Causality-Inspired Recipe for Generating Counterfactual Adversarial Examples''}

\bigskip

\section{Proof and Analysis}

\subsection{Proof of Equation~(2)}
\label{proof:linear_w}
\begin{proof}
For a linear regression model $\hat{\rv{y}}=\vecrv{w}^T\vecrv{x}$, where $\vecrv{x}=[\rv{x}_1,\rv{x}_2,\rv{x}_3]^T$, by Empirical Risk Minimization (ERM), we get the optimal solution:
\begin{equation*}
\begin{aligned}
\vecrv{w}&=(\vecscalar{X}^T\vecscalar{X})^{-1}\vecscalar{X}^T\vecscalar{y}=
(\sum_{i=1}^{N}{\vecscalar{x_i}\vecscalar{x_i}^T})^{-1}\sum_{i=1}^{N}{\vecscalar{x_i}y_i}\\
&=(\sum_{i=1}^{N}\left[
\begin{matrix} 
{x}_{i,1}^2 & {x}_{i,1}{x}_{i,2} & {x}_{i,1}{x}_{i,3}\\ 
{x}_{i,2}{x}_{i,1} & {x}_{i,2}^2 & {x}_{i,2}{x}_{i,3}\\
{x}_{i,3}{x}_{i,1} & {x}_{i,3}{x}_{i,2} & {x}_{i,3}^2\\
\end{matrix}\right])^{-1}
\sum_{i=1}^{N}\left[
\begin{matrix} 
{x}_{i,1} {y}_i\\ 
{x}_{i,2} {y}_i\\
{x}_{i,3} {y}_i\\
\end{matrix}\right]\\
&\overset{N\rightarrow \infty}{=}\mathbb{E}[\left[
\begin{matrix} 
\rv{x}_1^2 & \rv{x}_1\rv{x}_2 & \rv{x}_1\rv{x}_3\\ 
\rv{x}_2\rv{x}_1 & \rv{x}_2^2 & \rv{x}_2\rv{x}_3\\
\rv{x}_3\rv{x}_1 & \rv{x}_3\rv{x}_2 & \rv{x}_3^2\\
\end{matrix}\right]]^{-1}
\mathbb{E}[\left[
\begin{matrix} 
\rv{x}_1 \rv{y}\\ 
\rv{x}_2 \rv{y}\\
\rv{x}_3 \rv{y}\\
\end{matrix}\right]],
\end{aligned}
\end{equation*}
where $\vecscalar{X}\in\mathbb{R}^{N\times 3}$ is a matrix whose element $x_{i,j}$ denotes the $i$-th observation of $\rv{x}_j$, and $\vecscalar{y}\in\mathbb{R}^{N}$ is a vector whose element $y_i$ denotes the $i-th$ observation of $\rv{y}$.
According to the data-generating process in (1):
\begin{equation*}
\left\{
\begin{array}{ll}
      \rv{x}_1=\rv{u}_1 \quad &\rv{u}_1\sim\mathcal{N}(0, \sigma_{1}^2)  \\
      \rv{y} = a\rv{x}_1 + \rv{u}_y \quad &\rv{u}_y\sim\mathcal{N}(0, \sigma_{y}^2) \\
      \rv{x}_2 = b\rv{x}_3 + c\rv{y} + \rv{u}_2 \quad &\rv{u}_2\sim\mathcal{N}(0, \sigma_{2}^2) \\
      \rv{x}_3 = \rv{u}_3 \quad &\rv{u}_3\sim\mathcal{N}(0, \sigma_{3}^2)
\end{array}
,
\right.
\end{equation*}
we have:
\begin{equation*}
\begin{aligned}
\vecrv{w}&=\mathbb{E}[\left[
\begin{matrix} 
\rv{x}_1^2 & \rv{x}_1\rv{x}_2 & \rv{x}_1\rv{x}_3\\ 
\rv{x}_2\rv{x}_1 & \rv{x}_2^2 & \rv{x}_2\rv{x}_3\\
\rv{x}_3\rv{x}_1 & \rv{x}_3\rv{x}_2 & \rv{x}_3^2\\
\end{matrix}\right]]^{-1}
\mathbb{E}[\left[
\begin{matrix} 
\rv{x}_1 \rv{y}\\ 
\rv{x}_2 \rv{y}\\
\rv{x}_3 \rv{y}\\
\end{matrix}\right]]\\
&=\left[\begin{matrix}\sigma_{1}^{2} & \sigma_{1}^{2} a c & 0\\\sigma_{1}^{2} a c & \sigma_{1}^{2} a^{2} c^{2} + \sigma_{2}^{2} + \sigma_{3}^{2} b^{2} + \sigma_{y}^{2} c^{2} & \sigma_{3}^{2} b\\0 & \sigma_{3}^{2} b & \sigma_{3}^{2}\end{matrix}\right]^{-1}
\left[\begin{matrix}\sigma_{1}^{2} a\\\sigma_{1}^{2} a^{2} c + \sigma_{y}^{2} c\\0\end{matrix}\right]\\
&=\left[\begin{matrix}\frac{\sigma_{1}^{2} a^{2} c^{2} + \sigma_{2}^{2} + \sigma_{y}^{2} c^{2}}{\sigma_{1}^{2} \sigma_{2}^{2} + \sigma_{1}^{2} \sigma_{y}^{2} c^{2}} & - \frac{a c}{\sigma_{2}^{2} + \sigma_{y}^{2} c^{2}} & \frac{a b c}{\sigma_{2}^{2} + \sigma_{y}^{2} c^{2}}\\- \frac{a c}{\sigma_{2}^{2} + \sigma_{y}^{2} c^{2}} & \frac{1}{\sigma_{2}^{2} + \sigma_{y}^{2} c^{2}} & - \frac{b}{\sigma_{2}^{2} + \sigma_{y}^{2} c^{2}}\\\frac{a b c}{\sigma_{2}^{2} + \sigma_{y}^{2} c^{2}} & - \frac{b}{\sigma_{2}^{2} + \sigma_{y}^{2} c^{2}} & \frac{\sigma_{2}^{2} + \sigma_{3}^{2} b^{2} + \sigma_{y}^{2} c^{2}}{\sigma_{2}^{2} \sigma_{3}^{2} + \sigma_{3}^{2} \sigma_{y}^{2} c^{2}}\end{matrix}\right] 
\left[\begin{matrix}\sigma_{1}^{2} a\\\sigma_{1}^{2} a^{2} c + \sigma_{y}^{2} c\\0\end{matrix}\right]=\left[\begin{matrix}\frac{\sigma_{2}^{2} a}{\sigma_{2}^{2} + \sigma_{y}^{2} c^{2}}\\\frac{\sigma_{y}^{2} c}{\sigma_{2}^{2} + \sigma_{y}^{2} c^{2}}\\- \frac{\sigma_{y}^{2} b c}{\sigma_{2}^{2} + \sigma_{y}^{2} c^{2}}\end{matrix}\right].
\end{aligned}
\end{equation*}
\end{proof}

\subsection{Proof of Proposition~1}
\label{proof:yx_to_ymb}
\begin{proof}
Factorizing the conditional distribution via SCM $M$, we have:
\begin{equation*}
\begin{aligned}
p_{M}(\rv{y}|\vecrv{x})&=\frac{p_M(\vecrv{x},\rv{y})}{\int p_M(\vecrv{x},\rv{y})\, dy}\\
&=\frac{p_y(\rv{y}|\mathrm{Pa}_y)\prod_{\rv{x}_i\in \mathrm{Ch}_y^{M}}p_i(\rv{x}_i|\mathrm{Pa}_i^{M})\prod_{\rv{x}_i\in \vecrv{x} \setminus \mathrm{Ch}_y^{M}}p_i(\rv{x}_i|\mathrm{Pa}_i^{M})}{\int p_y(\rv{y}|\mathrm{Pa}_y^{M})\prod_{\rv{x}_i\in \mathrm{Ch}_y^{M}}p_i(\rv{x}_i|\mathrm{Pa}_i^{M}) \prod_{\rv{x}_i\in \vecrv{x} \setminus \mathrm{Ch}_y^{M}}p_i(\rv{x}_i|\mathrm{Pa}_i^{M}) \,dy}\\
&=\frac{p_y(\rv{y}|\mathrm{Pa}_y^{M})\prod_{\rv{x}_i\in \mathrm{Ch}_y^{M}}p_i(\rv{x}_i|\mathrm{Pa}_i^{M})}{\int p_y(\rv{y}|\mathrm{Pa}_y^{M})\prod_{\rv{x}_i\in \mathrm{Ch}_y^{M}}p_i(\rv{x}_i|\mathrm{Pa}_i^{M})\,dy}=p_M(\rv{y}|\mathrm{Mb}_y^{M}),
\end{aligned}
\end{equation*}
where $\mathrm{Pa}_i^{M}$, $\mathrm{Ch}_i^{M}$, and $\mathrm{Mb}_i^{M}$ denote the parents, children, and Markov blanket of $\rv{x}_i$ under the SCM $M$, respectively.
\end{proof}

\subsection{Proof of Proposition~2}
\label{proof:intervention_observe}
\begin{proof}
Given a SCM $M(\{\vecrv{x},\rv{y}\},f,\vecrv{u})$, the factorized conditional distribution $p_{M}(\rv{y}|\vecrv{x})$ is:
\begin{equation*}
p_{M}(\rv{y}|\vecrv{x})=\frac{p_y(\rv{y}|\mathrm{Pa}_y^{M})\prod_{\rv{x}_i\in \mathrm{Ch}_y^{M}}p_i(\rv{x}_i|\mathrm{Pa}_i^{M})}{\int p_y(\rv{y}|\mathrm{Pa}_y^{M})\prod_{\rv{x}_i\in \mathrm{Ch}_y^{M}}p_i(\rv{x}_i|\mathrm{Pa}_i^{M})\,dy}.
\end{equation*}

If an intervention on a \emph{child} variable of $\rv{y}$, $\rv{x}_j$, is conducted to obtain a new SCM $M'(\{\vecrv{x},\rv{y}\},f',\vecrv{u})$ with $f'=\{f_i: \rv{x}_i\neq\rv{x}_j\}\cup\{f'_j\}$, where $f_j'$ denotes the new mechanism of producing $\rv{x}_j$, then the interventional distribution:
\begin{equation*}
\begin{aligned}
p_{M'}(\rv{y}|\vecrv{x})&=\frac{p_{y}(\rv{y}|\mathrm{Pa}_y^{M'})\prod_{\rv{x}_i\in \mathrm{Ch}_y^{M'}}p_i(\rv{x}_i|\mathrm{Pa}_i^{M'})}{\int p_y(\rv{y}|\mathrm{Pa}_y^{M'})\prod_{\rv{x}_i\in \mathrm{Ch}_y^{M'}}p_i(\rv{x}_i|\mathrm{Pa}_i^{M'})\,dy}\\
&=\frac{p_y(\rv{y}|\mathrm{Pa}_y^{M})p_j(x_j)\prod_{\rv{x}_i\in \mathrm{Ch}_y^{M} \setminus \{\rv{x}_j\}}p_i(\rv{x}_i|\mathrm{Pa}_i^{M})}{\int p_y(\rv{y}|\mathrm{Pa}_y^{M})p_j(x_j)\prod_{\rv{x}_i\in \mathrm{Ch}_y^{M}\setminus \{\rv{x}_j\}}p_i(\rv{x}_i|\mathrm{Pa}_i^{M})\,dy}\\
&=\frac{p_y(\rv{y}|\mathrm{Pa}_y^{M})\prod_{\rv{x}_i\in \mathrm{Ch}_y^{M} \setminus \{\rv{x}_j\}}p_i(\rv{x}_i|\mathrm{Pa}_i^{M})}{\int p_y(\rv{y}|\mathrm{Pa}_y^{M})\prod_{\rv{x}_i\in \mathrm{Ch}_y^{M}\setminus \{\rv{x}_j\}}p_i(\rv{x}_i|\mathrm{Pa}_i^{M})\,dy}.
\end{aligned}
\end{equation*}

We prove the distribution $p_{M}(\rv{y}|\vecrv{x})\neq p_{M'}(\rv{y}|\vecrv{x})$ by contradiction. Suppose there exist a case where $p_{M}(\rv{y}|\vecrv{x})= p_{M'}(\rv{y}|\vecrv{x})$, then we have:
\begin{equation*}
\forall \vecrv{x}, \rv{y}, \quad \frac{p_{M}(\rv{y}|\vecrv{x})}{p_{M'}(\rv{y}|\vecrv{x})}=\frac{p_j(\rv{x}_j|\rv{y},\mathrm{Pa}_j^{M}\setminus\{\rv{y}\})\int p_y(\rv{y}|\mathrm{Pa}_y^{M})\prod_{\rv{x}_i\in \mathrm{Ch}_y^{M}\setminus \{\rv{x}_j\}}p_i(\rv{x}_i|\mathrm{Pa}_i^{M})\,dy}{\int p_y(\rv{y}|\mathrm{Pa}_y^{M})\prod_{\rv{x}_i\in \mathrm{Ch}_y^{M}}p_i(\rv{x}_i|\mathrm{Pa}_i^{M})\,dy}=1,
\end{equation*}
then we have:
\begin{equation*}
\forall \vecrv{x}, \rv{y}, \quad p_j(\rv{x}_j|\rv{y},\mathrm{Pa}_j^{M}\setminus\{\rv{y}\})=\frac{\int p_y(\rv{y}|\mathrm{Pa}_y^{M})\prod_{\rv{x}_i\in \mathrm{Ch}_y^{M}}p_i(\rv{x}_i|\mathrm{Pa}_i^{M})\,dy}{\int p_y(\rv{y}|\mathrm{Pa}_y^{M})\prod_{\rv{x}_i\in \mathrm{Ch}_y^{M}\setminus \{\rv{x}_j\}}p_i(\rv{x}_i|\mathrm{Pa}_i^{M})\,dy},
\end{equation*}
and for simplicity, we denote $h(\vecrv{x})=\frac{\int p_y(\rv{y}|\mathrm{Pa}_y^{M})\prod_{\rv{x}_i\in \mathrm{Ch}_y^{M}}p_i(\rv{x}_i|\mathrm{Pa}_i^{M})\,dy}{\int p_y(\rv{y}|\mathrm{Pa}_y^{M})\prod_{\rv{x}_i\in \mathrm{Ch}_y^{M}\setminus \{\rv{x}_j\}}p_i(\rv{x}_i|\mathrm{Pa}_i^{M})\,dy}$, since $\rv{y}$ is marginalized out, then we have:
\begin{equation*}
\forall \vecrv{x}, \rv{y}, \quad p_j(\rv{x}_j|\rv{y},\mathrm{Pa}_j^{M}\setminus\{\rv{y}\})=h(\vecrv{x}),
\end{equation*}
this will hold only when $\rv{y}$ makes no contribution to $p_j(\rv{x}_j|\rv{y},\mathrm{Pa}_j^{M}\setminus\{\rv{y}\})$, that is, $p_j(\rv{x}_j|\rv{y},\mathrm{Pa}_j^{M}\setminus\{\rv{y}\})=p_j(\rv{x}_j|\mathrm{Pa}_j^{M}\setminus\{\rv{y}\})$, which contradicts to our assumption of SCM $M$, i.e., $\rv{x}_j$ is the child of $\rv{y}$.
Thus we have $p_{M'}(\rv{y}|\vecrv{x})\neq p_{M}(\rv{y}|\vecrv{x})$.

If an intervention on a \emph{co-parent but not child} variable of $\rv{y}$, $\rv{x}_k$, is conducted to obtain a new SCM $M'(\{\vecrv{x},\rv{y}\},f',\vecrv{u})$ with $f'=\{f_i: \rv{x}_i\neq\rv{x}_k\}\cup\{f'_k\}$, where $f_k'$ denotes the new mechanism of producing $\rv{x}_k$, then the interventional distribution:
\begin{equation*}
\begin{aligned}
p_{M'}(\rv{y}|\vecrv{x})&=\frac{p_{y}(\rv{y}|\mathrm{Pa}_y^{M'})\prod_{\rv{x}_i\in \mathrm{Ch}_y^{M'}}p_i(\rv{x}_i|\mathrm{Pa}_i^{M'})}{\int p_y(\rv{y}|\mathrm{Pa}_y^{M'})\prod_{\rv{x}_i\in \mathrm{Ch}_y^{M'}}p_i(\rv{x}_i|\mathrm{Pa}_i^{M'})\,dy}\\
&\overset{\mathrm{Pa}_i^{M'}=\mathrm{Pa}_i^{M}}{=}\frac{p_y(\rv{y}|\mathrm{Pa}_y^{M})\prod_{\rv{x}_i\in \mathrm{Ch}_y^{M}}p_i(\rv{x}_i|\mathrm{Pa}_i^{M})}{\int p_y(\rv{y}|\mathrm{Pa}_y^{M})\prod_{\rv{x}_i\in \mathrm{Ch}_y^{M}}p_i(\rv{x}_i|\mathrm{Pa}_i^{M})\,dy},
\end{aligned}
\end{equation*}
thus we have  $p_{M'}(\rv{y}|\vecrv{x})=p_{M}(\rv{y}|\vecrv{x})$.
\end{proof}

\subsection{Proof and Analysis of Proposition~3}
\label{proof:latent}
\subsubsection{Proof of Proposition~3}
\begin{proof}
Given a SCM $M(\{\vecrv{x},\vecrv{z}\},\{f,g\},\{\vecrv{u}_x,\vecrv{u}_z\})$ where $f=\{f_i:\rv{z}_i\in\vecrv{z}\}$ with $\rv{z}_i=f_i(\mathrm{Pa}_i^M, \rv{u}_i)$ that parameterize the marginal $p_{f}(\vecrv{z})$, $\vecrv{x}=g(\vecrv{z},\vecrv{u}_x)$ that parameterize $p_{g}(\vecrv{x}|\vecrv{z})$, and target $\rv{y}$ is included in $\vecrv{z}$, i.e., $\vecrv{z}=[\rv{y},\rv{z}_1, ...,\rv{z}_n]^T$, the factorized conditional distribution:

\begin{equation*}
\begin{aligned}
p_{M}(\rv{y}|\vecrv{x})&=\frac{p_{M}(\vecrv{x}, y)}{p_{M}(\vecrv{x})}=\frac{\int p_{M}(\vecrv{x}, y, \rv{z}_1,...,\rv{z}_n)\, d(\rv{z}_1,...,\rv{z}_n)}{\int p_{M}(\vecrv{x}, \vecrv{z})\,d\vecrv{z}}\\
&=\frac{\int p_{g}(\vecrv{x}|\rv{y}, \rv{z}_1,...,\rv{z}_n)p_{f}(\rv{y},\rv{z}_1,...,\rv{z}_n)\, d(\rv{z}_1,...,\rv{z}_n)}{\int p_{g}(\vecrv{x}| \vecrv{z})p_{f}(\vecrv{z})\,d\vecrv{z}}
,
\end{aligned}
\end{equation*}
when interventions are conducted on the latent $\vecrv{z}$ to obtain a new SCM $M'(\{\vecrv{x},\vecrv{z}\},\{f',g\},\{\vecrv{u}_x,\vecrv{u}_z\})$, we have the conditional distribution under $M'$:
\begin{equation*}
\begin{aligned}
p_{M'}(\rv{y}|\vecrv{x})=\frac{\int p_{g}(\vecrv{x}|\rv{y}, \rv{z}_1,...,\rv{z}_n)p_{f'}(\rv{y},\rv{z}_1,...,\rv{z}_n)\, d(\rv{z}_1,...,\rv{z}_n)}{\int p_{g}(\vecrv{x}| \vecrv{z})p_{f'}(\vecrv{z})\,d\vecrv{z}},
\end{aligned}
\end{equation*}
note that since the generative process from $\vecrv{z}$ to $\vecrv{x}$, $g$ does not change, the distributions $p_{g}(\vecrv{x}| \vecrv{z})$ and $p_{g}(\vecrv{x}|\rv{y}, \rv{z}_1,...,\rv{z}_n)$ remain the same. If we have $p_{f'}(\vecrv{z})= p_{f}(\vecrv{z})$, then we have $p_{M'}(\rv{y}|\vecrv{x})= p_{M}(\rv{y}|\vecrv{x})$.
Thus, if we have $p_{M'}(\rv{y}|\vecrv{x})\neq p_{M}(\rv{y}|\vecrv{x})$, we will have $p_{f'}(\vecrv{z})\neq p_{f}(\vecrv{z})$.
\end{proof}

\subsubsection{Analysis of Proposition~3}
We prove that $p_{f'}(\vecrv{z})\neq p_{f}(\vecrv{z})$ is a necessary condition for $p_{M'}(\rv{y}|\vecrv{x})\neq p_{M}(\rv{y}|\vecrv{x})$, and here we further analyze how sufficient is the condition $p_{f'}(\vecrv{z})\neq p_{f}(\vecrv{z})$ for $p_{M'}(\rv{y}|\vecrv{x})\neq p_{M}(\rv{y}|\vecrv{x})$.
We start with investigating the case where the condition $p_{f'}(\vecrv{z})\neq p_{f}(\vecrv{z})$ can lead to $p_{M'}(\rv{y}|\vecrv{x})= p_{M}(\rv{y}|\vecrv{x})$. 
Suppose we have $p_{M'}(\rv{y}|\vecrv{x})= p_{M}(\rv{y}|\vecrv{x})$, then:
\begin{equation*}
\forall \vecrv{x}, \rv{y}, \quad \frac{\int p_{g}(\vecrv{x}|\rv{y}, \rv{z}_1,...,\rv{z}_n)p_{f'}(\rv{y},\rv{z}_1,...,\rv{z}_n)\, d(\rv{z}_1,...,\rv{z}_n)}{\int p_{g}(\vecrv{x}| \vecrv{z})p_{f'}(\vecrv{z})\,d\vecrv{z}}=\frac{\int p_{g}(\vecrv{x}|\rv{y}, \rv{z}_1,...,\rv{z}_n)p_{f}(\rv{y},\rv{z}_1,...,\rv{z}_n)\, d(\rv{z}_1,...,\rv{z}_n)}{\int p_{g}(\vecrv{x}| \vecrv{z})p_{f}(\vecrv{z})\,d\vecrv{z}}
\end{equation*}

Since we have $p_{f'}(\vecrv{z})\neq p_{f}(\vecrv{z})$ resulting in different joint distributions, i.e., $p_{M'}(\vecrv{x},\vecrv{z})\neq p_{M}(\vecrv{x},\vecrv{z})$, one case for $p_{M'}(\rv{y}|\vecrv{x})= p_{M}(\rv{y}|\vecrv{x})$ to hold is, the different joint distributions can results in the same marginals, that is, $\int p_{g}(\vecrv{x}|\rv{y}, \rv{z}_1,...,\rv{z}_n)p_{f}(\rv{y},\rv{z}_1,...,\rv{z}_n)\, d(\rv{z}_1,...,\rv{z}_n)=\int p_{g}(\vecrv{x}|\rv{y}, \rv{z}_1,...,\rv{z}_n)p_{f'}(\rv{y},\rv{z}_1,...,\rv{z}_n)\, d(\rv{z}_1,...,\rv{z}_n)$, $\int p_{g}(\vecrv{x}| \vecrv{z})p_{f}(\vecrv{z})\,d\vecrv{z}=\int p_{g}(\vecrv{x}| \vecrv{z})p_{f'}(\vecrv{z})\,d\vecrv{z}$, which is hardly holds and only hold in extreme case.
Besides, if the different joint distributions result in different marginals, another case for $p_{M'}(\rv{y}|\vecrv{x})= p_{M}(\rv{y}|\vecrv{x})$ to hold is, the effects of the different marginal distributions cancel each other out, which is a more extreme case.
In summary, $p_{f'}(\vecrv{z})\neq p_{f}(\vecrv{z})$ causing $p_{M'}(\rv{y}|\vecrv{x})= p_{M}(\rv{y}|\vecrv{x})$ only occur in the aforementioned two special cases, and if the parameters of the generated mechanism is randomly specify, then the probability of such special occurrence should have measure zero.
Therefore, at most cases, if $p_{f'}(\vecrv{z})\neq p_{f}(\vecrv{z})$, then  $p_{M'}(\rv{y}|\vecrv{x})\neq p_{M}(\rv{y}|\vecrv{x})$.

\section{Detailed Experimental Setup}
\label{app:inplementations}

\subsection{Detailed Dataset Preprocessing}
\begin{figure}[h]
    \centering
    \includegraphics[scale=0.6]{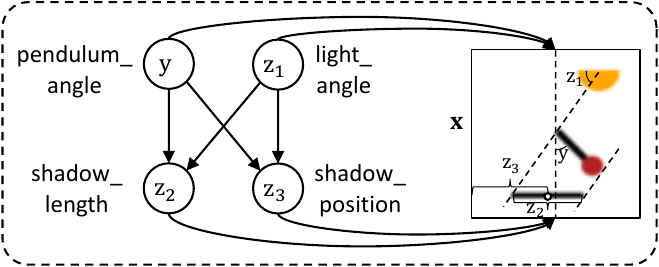}
    \caption{The generating process of Pendulum.}
    \label{fig:dgp_pend}
\end{figure}

\label{app:dataset}
\subsubsection{Pendulum}
In our experiments, we adopt a synthetic dataset \emph{Pendulum} proposed by \cite{c:causalvae} which consists of four causal related factors, i.e., \emph{pendulum\_angle}, \emph{light\_angle}, \emph{shadow\_length}, and \emph{shodow\_position}, where the generative process is depicted in Figure~\ref{fig:dgp_pend}. Given the \emph{pendulum\_angle} and \emph{light\_angle}, we can determined the \emph{shadow\_length} and \emph{shadow\_position}, following the projection law.
In detail, the ground truth generative process of the Pendulum dataset is shown as follows.
\begin{equation*}
\left\{
\begin{array}{ll}
\rv{y} =\rv{u}_y \quad \rv{u}_y\sim U(0,\frac{\pi}{4}) \\
\rv{z}_1 =\rv{u}_1 \quad \rv{u}_1\sim U(\frac{\pi}{4},\frac{\pi}{2}) \\
\rv{z}_2 = (c_x+l_p\mathop{\sin}\rv{y}-\frac{c_y-l_p\mathop{\cos}\rv{y}-b}{\mathop{\tan}\rv{z}_1})-(c_x-\frac{c_y-b}{\mathop{\tan}\rv{z}_1}) \quad \\
\rv{z}_3 = (c_x+l_p\mathop{\sin}\rv{y}-\frac{c_y-l_p\mathop{\cos}\rv{y}-b}{\mathop{\tan}\rv{z}_1}+c_x-\frac{c_y-b}{\mathop{\tan}\rv{z}_1}) / 2
\end{array}
\right.
\end{equation*}
where $c_x=10$, $c_y=10.5$ are constants denoting the axis of the center, $l_p=9.5$ denotes the pendulum length including the red ball, the bottom line of a single image corresponding to $y=b$ with $b=-0.5$. 
To make the dataset realistic, we further introduce random noises on \emph{pendulum\_angle} on 15~\% of data when generating images, representing measurement error.
The Pendulum generator then synthesizes images according to the above generative process of size $96\times96$ with 4 channels.
To train the victim models and the causal generative model, we pre-process the image by resizing it to $64\times64$, then converting it to RGB space with 3 channels.
Specifically, for the victim model, our task is to predict the \emph{pendulum\_angle}. We convert the continuous \emph{pendulum\_angle} to 50 categorical labels based on which angle intervals an image belongs, to achieve the image classification task.

\subsubsection{CelebA}
The dataset contains 20K human face images with 40 labeled binary attributes. In our experiments, we adopt the causal graph called CelabA(Attractive) (depicted in Figure~3~(b)) proposed in \cite{a:dear} as our causal knowledge to validate our CADE and train the causal generative model. 
To train the victim model and causal generative model, we pre-process the image by, taking crops of $128\times128$ then resizing it to $64\times64$ resolution with 3 channels. 
Specifically, the task of the victim classification models is to predict if a person is \emph{young} or not given a face image.

\subsubsection{SynMeasurement}
The synthesized dataset contains 8 endogenous including the target $\rv{y}$ where the data-generating process is shown as follows:
\begin{equation*}
\left\{
\begin{array}{ll}
\rv{x}_{A} =\rv{u}_{A} \quad &\rv{u}_{A}\sim \mathcal{N}(0,1) \\
\rv{x}_{P1} =\rv{x}_{A}+\rv{u}_{P1} \quad &\rv{u}_{P1}\sim \mathcal{N}(0,1) \\
\rv{x}_{P2} =\rv{x}_{A}+\rv{u}_{P2} \quad &\rv{u}_{P2}\sim \mathcal{N}(0,1) \\
\rv{y} =\rv{x}_{P1}+\rv{x}_{P2}+\rv{u}_{y} \quad &\rv{u}_{y}\sim \mathcal{N}(0,1) \\
\rv{x}_{CP} =\rv{u}_{CP} \quad &\rv{u}_{CP}\sim \mathcal{N}(0,1) \\
\rv{x}_{C1} =\rv{x}_{CP}+4\rv{y} + \rv{u}_{C1} \quad &\rv{u}_{C1}\sim \mathcal{N}(0,1) \\
\rv{x}_{C2} =\rv{x}_{C1}+\rv{y} + \rv{u}_{C2} \quad &\rv{u}_{C2}\sim \mathcal{N}(0,1) \\
\rv{x}_{D} =\rv{x}_{C1}+\rv{x}_{C2} + \rv{u}_{D} \quad &\rv{u}_{D}\sim \mathcal{N}(0,1) \\

\end{array}
\right.
\end{equation*}

As in our experiments, the victim models use $\vecrv{x}=[\rv{x}_{A},\rv{x}_{P1},\rv{x}_{P2},\rv{x}_{CP},\rv{x}_{C1},\rv{x}_{C2},\rv{x}_{D}]^T$ to predict the target $\rv{y}$ for a regression task.

\subsection{Latent Intervention for Pendulum and CelebA}
\label{app:setup_latent}
To generate adversarial examples for two image datasets, Pendulum and CelabA, we leverage a state-of-the-art causal generative model, DEAR \cite{a:dear}, which can generate interventional examples by intervention on the latent. More specifically, DEAR consists of an encoder, decoder, and a causal layer that parameterizes the causal generating process of the latent. To model this generating process, we adopt a general non-linear SCM proposed in \cite{c:daggnn}, which is:
\begin{equation}
\label{eq:nlr_scm}
f(\vecrv{z}) = \vecrv{A}^Tf(\vecrv{z})+\vecrv{u}_z,
\end{equation}
where $f$ denotes an invertible non-linear transformation.
To generate an adversarial example of an image $\vecrv{x}$, the first step is to recover the causal-related latent codes through the encoder, i.e., $\vecrv{z}=Encoder(\vecrv{x})$. Next, interventions are conducted in the latent space to generate counterfactual adversarial $\vecrv{z}^{adv}$, where the first step is to recover the exogenous that encodes the individual information of the current world:
\begin{equation}
\label{eq:abduction_latent}
\vecrv{u}_z = (\vecrv{I}-\vecrv{A}^T)f(\vecrv{z}),
\end{equation}
then interventions on some desired variables $\vecrv{z}_S$ are conducted to obtain $\vecrv{z}'$, then predict the consequences of such intervention to obtain $\vecrv{z}^{adv}$, which is:
\begin{equation}
\label{eq:act_pre_latent}
\vecrv{z}^{adv}=f^{-1}(\vecrv{A}^Tf(\vecrv{z}')\odot (1-\vecrv{m})+f(\vecrv{z}')\odot \vecrv{m}+ \vecrv{u}_z\odot (1-\vecrv{m})),
\end{equation}
by iterate the \eqref{eq:act_pre_latent} $l$ times where $l$ denotes the depth of the causal graph, we obtain $\vecrv{z}^{adv}$. The final step is to predict the consequence of the total intervention on $\vecrv{z}$ to update $\vecrv{x}$, by feeding $\vecrv{z}^{adv}$ to the decoder of the causal generative model, which is, $\vecrv{x}^{adv}=Decoder(\vecrv{z}^{adv})$.
For $f$, we use piece-wise linear functions to model the element-wise non-linearity as in \cite{a:dear}, since the family of such piece-wise linear functions is expressive enough to model general element-wise non-linear invertible transformations.

\subsection{The CADE Algorithm}
\label{app:algorithm}
We showcase the algorithms of our CADE for the observable and latent variables in the following Algorithms~\ref{alg:cade} and \ref{alg:cade_latent}, respectively.
 \begin{algorithm}
\caption{The CADE algorithm for the observable}\label{alg:cade}
\begin{algorithmic}
\Require Original example $\vecrv{x},\rv{y}$, the substitute model $f_{\theta}$, $(\vecrv{A},f)$ that parameterize the generative process, the depth of causal graph $l$, intervened indices $S$, intervention budget $\epsilon$, step\_size $\alpha$, num\_steps $n$.
\Ensure The adversarial example $\vecrv{x}^{adv}$.
\State Initialize $\vecrv{x}^{adv, (0)}\gets\vecrv{x}$
\State $\vecrv{u}\gets$ Recover the exogenous by Equation~(4).
\For{$i=1,2,...,n$}
\State $\vecrv{x}'^{(i)}\gets \vecrv{x}'^{(i-1)}- \alpha \frac{-\partial\mathcal{L}_{pred}(f_{\theta}(\vecrv{x}^{adv, (i-1)}),\rv{y})}{\partial\vecrv{x}^{adv,(i-1)}_S}$ (Incorporate with Adam optimizer in our implementation)
\State $\Delta\vecrv{x}^{(i)}\gets\vecrv{x}'^{(i)}-\vecrv{x}$
\State $\Delta\vecrv{x}^{(i)}\gets$ Clamp $\Delta\vecrv{x}^{(i)}$ such that $\lVert \Delta\vecrv{x}\rVert_{p} \leq \epsilon$
\State $\vecrv{x}'^{(i)}\gets\Delta\vecrv{x}^{(i)}+\vecrv{x}$
\For{$j=1,2...,l$}
\State $\vecrv{x}^{adv, (i)}\gets$ Update the consequences by Equation~(5).
\State $\vecrv{x}'^{(i)} \gets \vecrv{x}^{adv, (i)}$
\EndFor
\EndFor
\State $\vecrv{x}^{adv}\gets\vecrv{x}^{adv, (n)}$.
\end{algorithmic}
\end{algorithm}

\begin{algorithm}
\caption{The CADE algorithm for the latent}\label{alg:cade_latent}
\begin{algorithmic}
\Require Original example $\vecrv{x},\rv{y}$, the substitute model $f_{\theta}$, $(\vecrv{A},f,Encoder,Decoder)$ that parameterize the generative process, the depth of causal graph $l$, intervened indices $S$, intervention budget $\epsilon$, step\_size $\alpha$, num\_steps $n$.
\Ensure The adversarial example $\vecrv{x}^{adv}$.
\State $\vecrv{z} \gets Encoder(\vecrv{x})$
\State Initialize $\vecrv{z}^{adv, (0)}\gets\vecrv{z},\vecrv{x}^{adv, (0)}\gets Decoder(\vecrv{z})$
\State $\vecrv{u}_{z}\gets$ Recover the exogenous by Equation~\eqref{eq:abduction_latent}.
\For{$i=1,2,...,n$}
\State $\vecrv{z}'^{(i)}\gets \vecrv{z}'^{(i-1)}- \alpha \frac{-\partial\mathcal{L}_{pred}(f_{\theta}(\vecrv{x}^{adv, (i-1)}),\rv{y})}{\partial\vecrv{z}^{adv,(i-1)}_S}$ (Incorporate with Adam optimizer in our implementation)
\State $\Delta\vecrv{z}^{(i)}\gets\vecrv{z}'^{(i)}-\vecrv{z}$
\State $\Delta\vecrv{z}^{(i)}\gets$ Clamp $\Delta\vecrv{z}^{(i)}$ such that $\lVert \Delta\vecrv{z}\rVert_{p} \leq \epsilon$
\State $\vecrv{z}'^{(i)}\gets\Delta\vecrv{z}^{(i)}+\vecrv{z}$
\For{$j=1,2...,l$}
\State $\vecrv{z}^{adv, (i)}\gets$ Update the consequences by Equation~\eqref{eq:act_pre_latent}.
\State $\vecrv{z}'^{(i)} \gets \vecrv{z}^{adv, (i)}$
\EndFor
\State $\vecrv{x}^{adv, (i)} \gets Decoder(\vecrv{z}^{adv, (i)})$
\EndFor
\State $\vecrv{x}^{adv}\gets\vecrv{x}^{adv, (n)}$.
\end{algorithmic}
\end{algorithm}

\subsection{Implementation Detail}

\paragraph{Victim Model}
We select the following network architectures as our victim models:
\begin{itemize}
    \item \textbf{Image classification}: Res-50 \cite{c:resnet}, VGG-16 \cite{c:vgg} and their adversarial trained defense variants, Res-50(D) and VGG-16(D). We select the PGD algorithm \cite{c:pgd} ($\epsilon$ = 8 / 255 for Pendulum and 4 / 255 for CelebA, num\_steps = 10) to adversarial train our defense models.
    \item \textbf{Measurement regression}: Linear ([dim\_input, 1]) and MLP ([dim\_input, 32, 1]), and their adversarial trained variants, Linear(D) and MLP(D) incorporated with PGD ($\epsilon$ = 0.1, num\_steps = 5).
\end{itemize}

\paragraph{Causal Generative model for Pendulum and CelebA}
We adopt DEAR as our causal generative model, and train the model using the hyper-parameter settings proposed in \cite{a:dear} for both Pendulum and CelebA. 
To generate more imperceptible examples, we further fine-tune the encoder structure by minimizing a reconstruction loss after the training.

\paragraph{Hyper-Parameter Setting of Attacks}
We adopt several baselines for comparisons on Pendulum and CelebA, where the hyper-parameter settings of each approach are:
\begin{itemize}
    \item  \textbf{PGD} \cite{c:pgd}: perturbation budget $\epsilon$: 8 / 255 (with metrics $\mathcal{L}_{\inf}$), num\_steps: 10, step\_size: 0.01 for Pendulum and 0.05 for CelebA.
    \item \textbf{C\&W} \cite{c:cw}: c: 1 (penalize the prediction loss), kappa: 0, distance loss: $\mathcal{L}_2$, num\_steps: 50, step\_size: 0.01.
    \item  \textbf{SAE} \cite{c:sae}: num\_steps: 500.
    \item \textbf{ACE} \cite{c:ace}: num\_steps: 100, step\_size: 0.01, num\_pieces K: 64.
    \item  \textbf{APGD($\mathcal{L}_{\inf}$)} \cite{c:apgd}: perturbation budget $\epsilon$: 8 / 255 (with metrics $\mathcal{L}_{\inf}$), num\_steps: 5.
    \item  \textbf{APGD($\mathcal{L}_{1}$)} \cite{c:apgdl1}: perturbation budget $\epsilon$: 12.0 (with metrics $\mathcal{L}_{1}$), num\_steps: 25 for Pendulum and 100 for CelebA.
    \item  \textbf{CADE} (Ours): intervention budget $\epsilon$ (latent space): $0.3\times range$ ($range_i=\rv{z}_i^{max}-\rv{z}_i^{min}$ for each $i$) for Pendulum and $0.7\times range$ for CelebA with metrics $\mathcal{L}_{\inf}$, num\_steps: 20 for Pendulum and 200 for CelebA, step\_size: 0.4 for Pendulum and 0.5 for CelebA.
    \item  \textbf{CADE} (Ours, on SynMeasurement Experiment): intervention budget $\epsilon$: $[0.05, 0.1, 0.15, 0.2, 0.25, 0.3, 0.35, 0.4, 0.45, 0.5] \times range$ with metrics $\mathcal{L}_{\inf}$, num\_steps: 150, step\_size: 0.1.
\end{itemize}

\section{Additional Experimental Result}

\subsection{The Sensitivity of the Intervention Budget}

\begin{figure}
    \centering
    \includegraphics[scale=0.5]{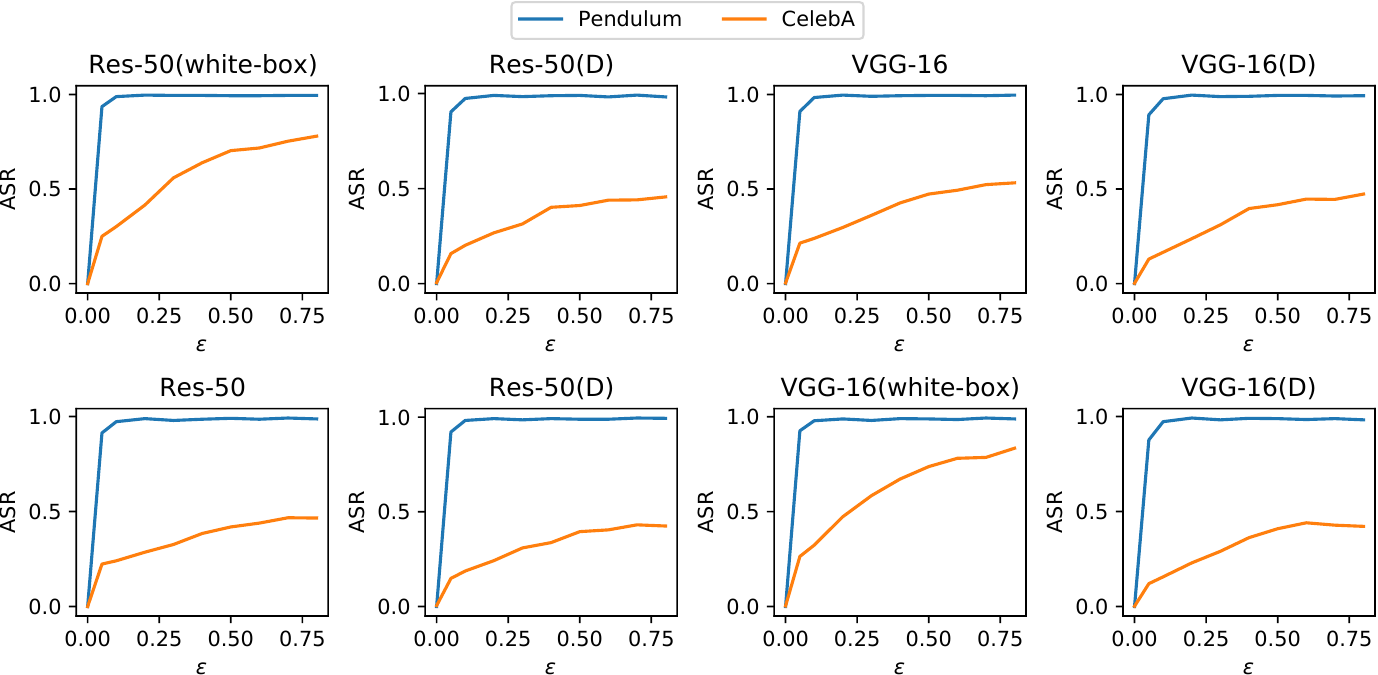}
    \caption{Attack Success Rate (ASR) w.r.t. intervention budgets $\epsilon$ of CADE on Pendulum and CelebA.} 
    \label{fig:res_sensitivity}
\end{figure}

The intervention budget $\epsilon$ as a hyper-parameter can control the upper bound of the intervention magnitudes, which is an indicator of the size of the search space. 
In CADE, when attacking images, the $\epsilon$ is applied in the latent space to constraint $\vecrv{z}$. Here, we showcase the attack success rates of our CADE with different $\epsilon$ on both Pendulum and CelebA datasets in Figure~\ref{fig:res_sensitivity}. 
Specifically, for Pendulum, our CADE intervenes on $\vecrv{z}_{1:3}$, where the corresponding causal graph is shown in Figure~1~(a), and for CelabA, our CADE intervenes on $\vecrv{z}_{1:5}$, where the corresponding causal graph is shown in Figure~1~(b).
From Figure~\ref{fig:res_sensitivity}, we observe the increasing attack success rates with the increasing intervention budget $\epsilon$ on Pendulum and CelebA, indicating the interventional data drawn from a more extended space has a greater capacity to flip the prediction.

\begin{figure}
    \centering
    \includegraphics[scale=0.5]{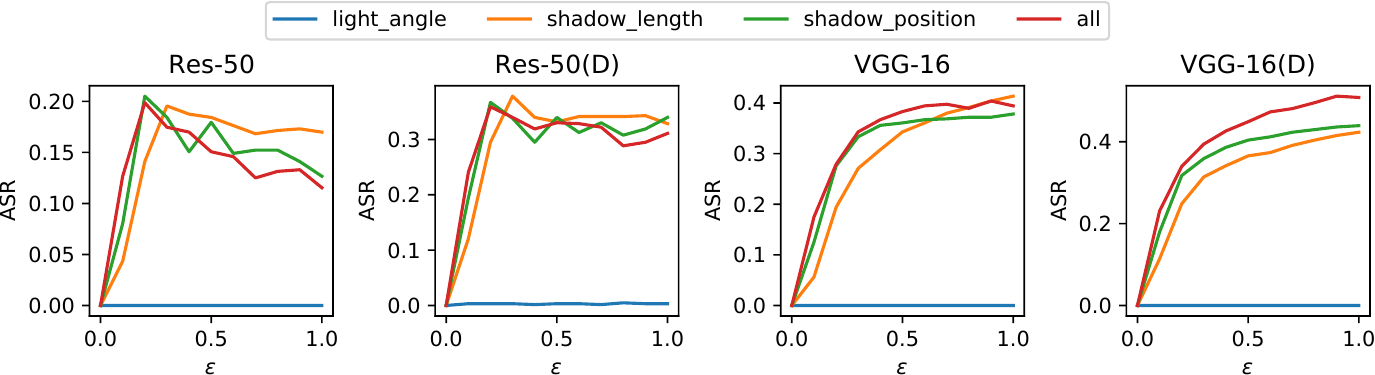}
    \caption{Attack Success Rate (ASR) w.r.t. intervention budgets $\epsilon$ of simulator attack on Pendulum.} 
    \label{fig:res_sim}
\end{figure}

\subsection{Attack on Pendulum by Simulator}
Instead of generating adversarial examples by generative model conditioning on specific latent codes, we further investigate the attack performance on \emph{Pendulum} dataset with adversarial examples generated by the \emph{Pendulum} simulator, which can generate perfect images without any noises. 
To perform the interventions, we first add noises $\rv{\delta z}_i$ uniformly sampled from $\rv{\delta z}_i \sim U(-\epsilon \times range_i, \epsilon \times range_i)$ to each latent variable $\rv{z}_i$ to obtain $\rv{z}_{i}^{adv}$, i.e., $\rv{z}_{i}^{adv}=\rv{z}_{i}+\rv{\delta z}_i$, then leverage the \emph{Pendulum} simulator to generate the counterfactual adversarial image $\vecrv{x}^{adv}=g_{simulator}(\vecrv{z}^{adv})$, where $\epsilon$ denotes the intervention budget that controls the upper bound of the intervention magnitude, and $range_i$ denotes the range of each latent variable $\rv{z}_i$. 
Specifically, to obtain valid latent $\rv{z}_{i}^{adv}$ for image generation, we clip $\rv{z}_{i}^{adv}$ by its valid minimum and maximum values, i.e., $\rv{z}_i^{adv}=\min(\max(\rv{z}_i^{adv}, \rv{z}_i^{min}), \rv{z}_i^{max})$, where $\rv{z}_i^{min}$ and $\rv{z}_i^{max}$ denote the minimum and maximum of $\rv{z}_i$, respectively.
In our experiment, we investigate the intervention on \emph{light\_angle} (co-parent of $\rv{y}$), \emph{shadow\_length} (child of $\rv{y}$), \emph{shadow\_position} (child of $\rv{y}$), and all the three variables (shortly denoted as \emph{all}), and the result is shown in Figure~\ref{fig:res_sim}. 
From Figure~\ref{fig:res_sim} we observe that intervention on \emph{light\_angle} (co-parent of $\rv{y}$) is not effective by achieving nearly zero ASR, while the other three are effective in attacking. Proposition~3 can suggest the results, that is, adversarial examples generated from \emph{light\_angle} intervention are sampled from $\vecrv{z}$ without any structural changes. In contrast, the adversarial examples generated by the other three interventions are sampled from different marginals $p_{f'}(\vecrv{z})$ with structural changes in $\vecrv{z}$. 
Moreover, compared with the results by CADE using the generative model in Table~1, we observe that CADE with generative model yields higher ASR than the simulator attack, even in the \emph{light\_angle} intervention. This can be due to: 1) CADE with generative model leverages the gradient information when attacking to further exploit the weakness of the classifier model; 2) The classifier model that overfits to the perfect image tends to make mistakes when the imperfect legitimate examples are generated from the generative model with noises, which also can be suggested in \cite{c:acgan}.

\subsection{Additional Case Study}
Here we showcase more qualitative case studies in the following Figures~\ref{fig:quali_app_pend} and \ref{fig:quali_app_celeba}.

\begin{figure}
    \centering
    \includegraphics[scale=0.42]{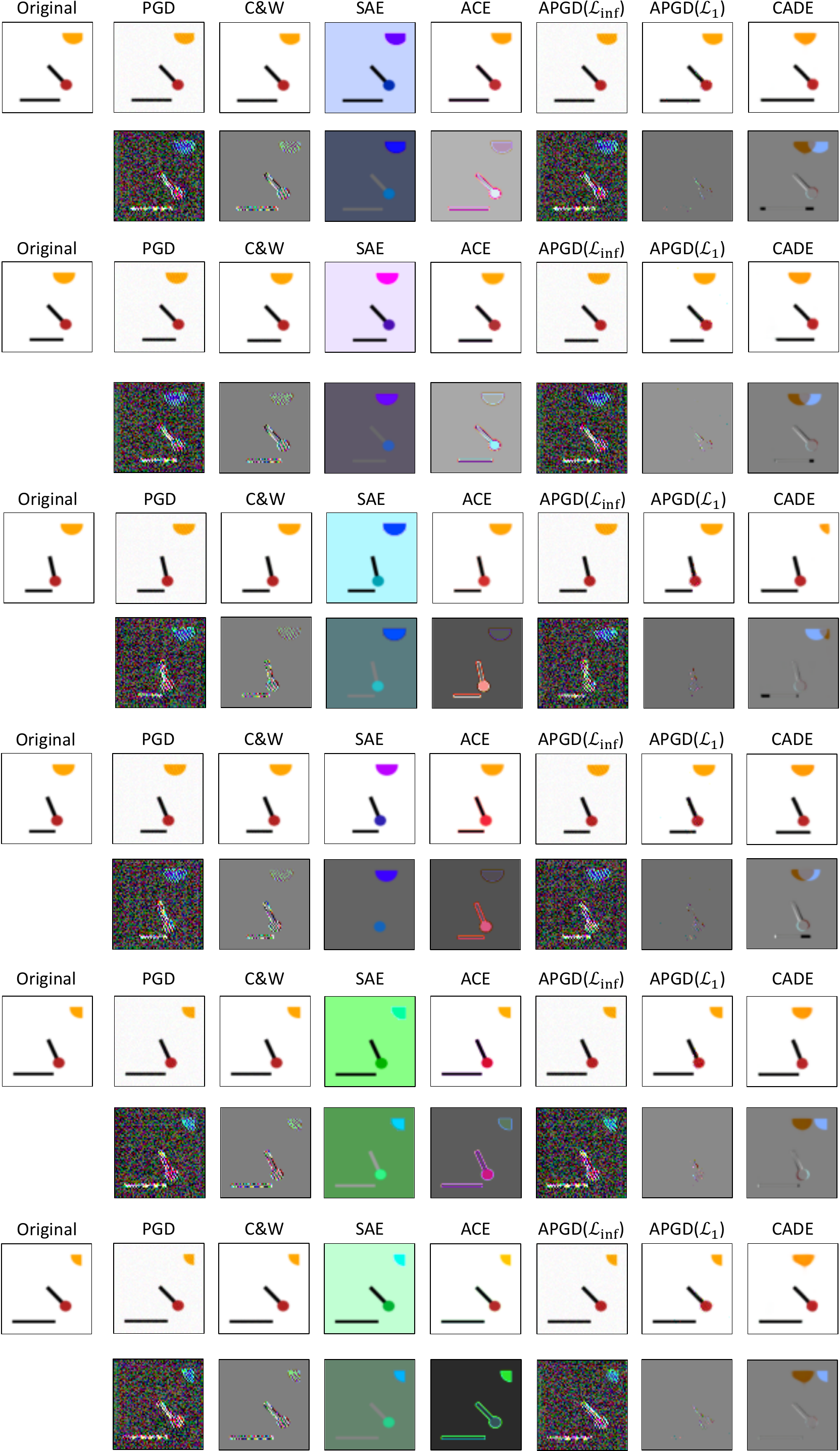}
    \caption{Visualization of adversarial examples on Pendulum obtained by different approaches.}
    \label{fig:quali_app_pend}
\end{figure}

\begin{figure}
    \centering
    \includegraphics[scale=0.42]{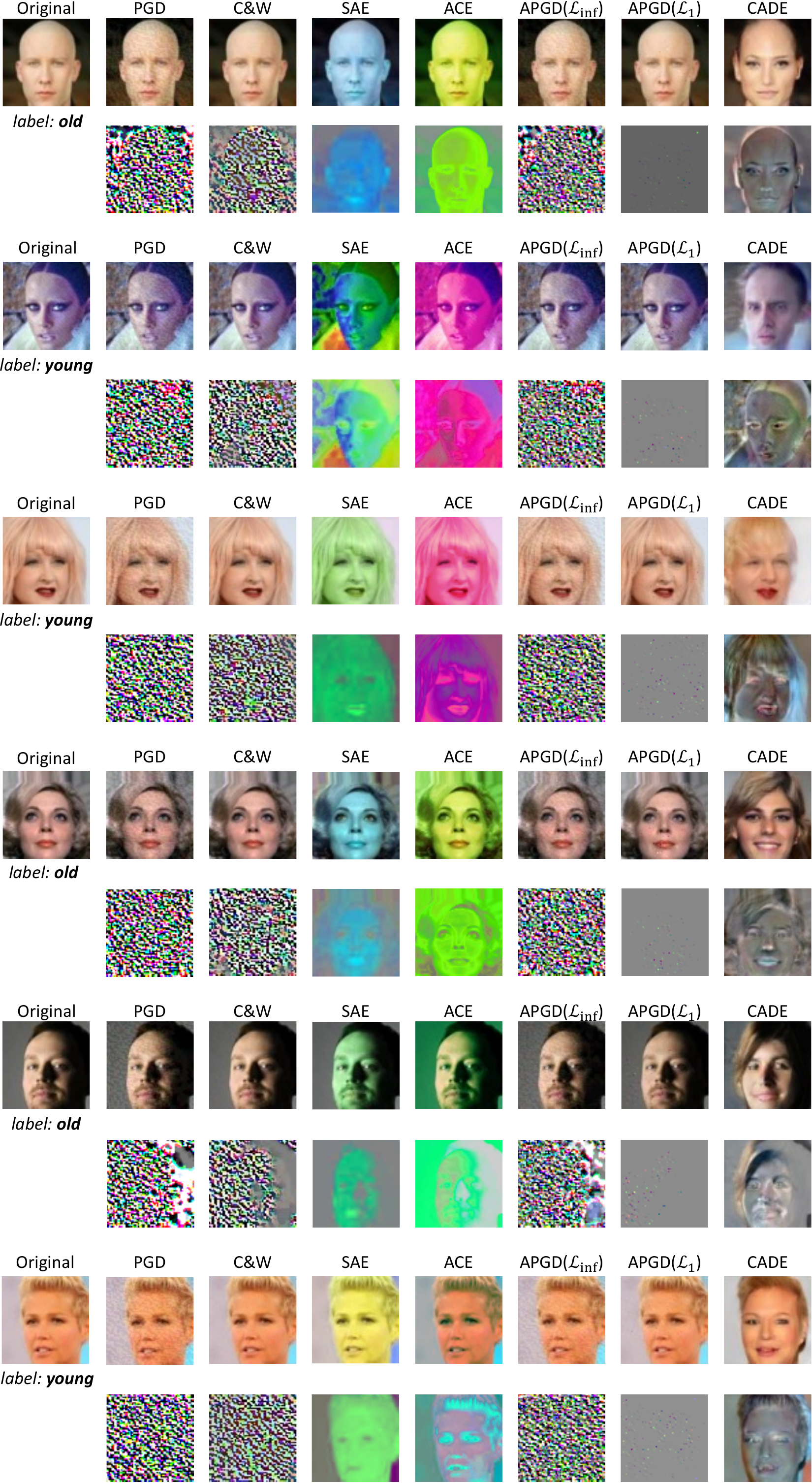}
    \caption{Visualization of adversarial examples on CelebA obtained by different approaches.}
    \label{fig:quali_app_celeba}
\end{figure}

\end{document}